\documentclass{article}

\PassOptionsToPackage{numbers}{natbib}
\usepackage[main,final]{neurips_2026}

\usepackage[utf8]{inputenc} %
\usepackage[T1]{fontenc}    %
\usepackage{hyperref}       %
\usepackage{url}            %
\usepackage{booktabs}       %
\usepackage{amsfonts}       %
\usepackage{nicefrac}       %
\usepackage{microtype}      %
\usepackage[dvipsnames]{xcolor} %

\usepackage{microtype}
\usepackage{graphicx}
\usepackage{subcaption}
\usepackage{hyperref}
\usepackage{enumitem}
\usepackage{multirow}

\usepackage{amsmath}
\usepackage{amssymb}
\usepackage{mathtools}
\usepackage{amsthm}
\usepackage{pifont}
\usepackage[capitalize,noabbrev]{cleveref}
\usepackage{tikz}
\usepackage{colortbl}
\usepackage[capitalize,noabbrev]{cleveref}

\theoremstyle{plain}

\theoremstyle{definition}

\theoremstyle{remark}

\newcommand{\revise}[1]{\textcolor{black}{#1}}
\newcommand{\githubicon}{%
  \raisebox{-1.5pt}{\includegraphics[height=1.05em]{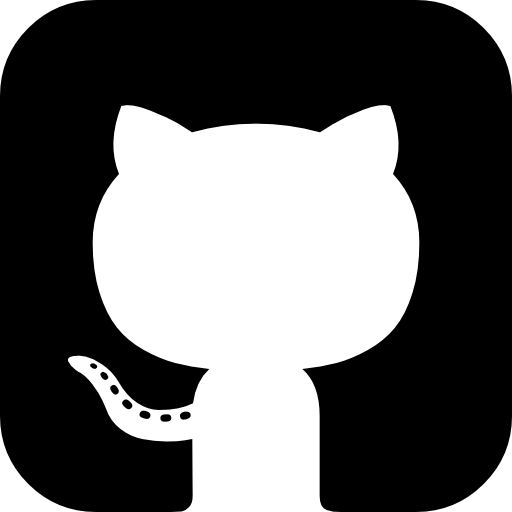}}%
}
\newcommand{\huggingfaceicon}{%
  \raisebox{-1.5pt}{\includegraphics[height=1.05em]{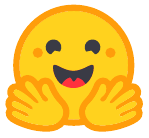}}%
}

\definecolor{linkColor}{rgb}{0.2,0.4,0.6}
\definecolor{myblue}{HTML}{0379AC}
\definecolor{myred}{HTML}{A50E50}
\definecolor{myorange}{RGB}{238, 133, 74}
\definecolor{latentcolor}{named}{cyan}
\definecolor{normalcolor}{RGB}{0, 0, 0}
\usepackage{marvosym}
\definecolor{lightblue1}{rgb}{0.97, 0.985, 1} 
\definecolor{lightblue2}{rgb}{0.92, 0.965, 1} 
\definecolor{lightblue3}{rgb}{0.84, 0.93, 1}
\definecolor{lightblue4}{rgb}{0.74, 0.87, 1}
\definecolor{lightblue5}{rgb}{0.64, 0.81, 1}
\definecolor{lightblue6}{rgb}{0.54, 0.75, 1}

\definecolor{lightgreen1}{rgb}{0.97, 1.00, 0.97}
\definecolor{lightgreen2}{rgb}{0.92, 0.98, 0.92}
\definecolor{lightgreen3}{rgb}{0.84, 0.95, 0.84}
\definecolor{lightgreen4}{rgb}{0.74, 0.91, 0.74}
\definecolor{lightgreen5}{rgb}{0.64, 0.86, 0.64}
\definecolor{lightgreen6}{rgb}{0.54, 0.81, 0.54}

\definecolor{lightorange1}{rgb}{1.00, 0.98, 0.95}
\definecolor{lightorange2}{rgb}{1.00, 0.95, 0.85}
\definecolor{lightorange3}{rgb}{1.00, 0.90, 0.70}
\definecolor{lightorange4}{rgb}{1.00, 0.85, 0.55}
\definecolor{lightorange5}{rgb}{1.00, 0.80, 0.40}
\definecolor{lightorange6}{rgb}{1.00, 0.75, 0.30}

\definecolor{lightpurple1}{rgb}{0.985, 0.97, 1.00}
\definecolor{lightpurple2}{rgb}{0.96, 0.92, 1.00}
\definecolor{lightpurple3}{rgb}{0.93, 0.84, 1.00}
\definecolor{lightpurple4}{rgb}{0.87, 0.74, 1.00}
\definecolor{lightpurple5}{rgb}{0.81, 0.64, 1.00}
\definecolor{lightpurple6}{rgb}{0.75, 0.54, 1.00}

\definecolor{lightred1}{rgb}{1.00, 0.97, 0.97}
\definecolor{lightred2}{rgb}{1.00, 0.92, 0.92}
\definecolor{lightred3}{rgb}{1.00, 0.84, 0.84}
\definecolor{lightred4}{rgb}{1.00, 0.74, 0.74}
\definecolor{lightred5}{rgb}{1.00, 0.64, 0.64}
\definecolor{lightred6}{rgb}{1.00, 0.54, 0.54}

\definecolor{lightcyan1}{rgb}{0.97, 1.00, 1.00}
\definecolor{lightcyan2}{rgb}{0.92, 0.98, 0.98}
\definecolor{lightcyan3}{rgb}{0.84, 0.95, 0.96}
\definecolor{lightcyan4}{rgb}{0.74, 0.91, 0.94}
\definecolor{lightcyan5}{rgb}{0.64, 0.87, 0.92}
\definecolor{lightcyan6}{rgb}{0.54, 0.83, 0.90}
\usepackage{colortbl}
\definecolor{Gray}{gray}{0.85}
\definecolor{LightCyan}{rgb}{0.88,1,1}
\definecolor{greyC}{RGB}{180,180,180}
\definecolor{greyL}{RGB}{235,235,235}
\definecolor{citeColor}{RGB}{0,20,115}
\usepackage{hyperref}
\usepackage{tcolorbox}
\hypersetup{colorlinks,linkcolor={red},citecolor={citeColor},urlcolor={citeColor}}
\definecolor{shadecolor}{rgb}{0.92,0.92,0.92}
\usepackage[textsize=tiny]{todonotes}
\newcommand{\method}{\textit{Composition-RL}}
\newcommand{\up}[1]{\textcolor{OliveGreen}{\small \ $\uparrow${#1}}}
\newcommand{\down}[1]{\textcolor{Maroon}{\small \ $\downarrow${#1}}}
\newcommand{\datamethod}{\textit{SPC}}
\newcommand{\solveall}{\textit{solve\_all}}
\newcommand{\solvenone}{\textit{solve\_none}}
\newcommand{\compose}{\textit{Compose}}
\newcommand{\dataset}{\texttt{MATH-Composition-199K}}
\newcommand{\datasetphysicsmath}{\texttt{Physics-MATH-Composition-141K}}

\title{\method{}: Compose Your Verifiable Prompts for Reinforcement Learning of Large Language Models}

\author{%
  \bfseries\small
  Xin Xu$^{1,2}$, Clive Bai$^{2}$, Kai Yang$^{2}$, Tianhao Chen$^{1}$,
  Cai Yuchen$^{2}$, Yangkun Chen$^{2}$, \\
  \bfseries\small
  Weijie Liu$^{2}$, Hao Chen$^{6}$, Yang Wang$^{5}$,
  Saiyong Yang$^{2,\dagger}$, Can Yang$^{1,3,4,7,\dagger}$ \\
  \normalfont\scriptsize
  $^{1}$Department of Mathematics, HKUST \enspace $^{2}$Hy, Tencent \enspace
  $^{3}$Big Data Bio-Intelligence Lab, HKUST \\[-0.15em]
  \normalfont\scriptsize
  $^{4}$State Key Laboratory of Nervous System Disorders, HKUST \enspace
  $^{5}$Department of Mathematics, HKU \\[-0.15em]
  \normalfont\scriptsize
  $^{6}$Department of Computer Science and Engineering, HKUST \enspace
  $^{7}$IAS Center for AI for Scientific Discoveries, HKUST \\
  \normalfont\footnotesize
  \githubicon\,
  \href{https://github.com/XinXU-USTC/Composition-RL}{\textbf{GitHub Repository}} \quad
  \huggingfaceicon\,
  \href{https://huggingface.co/collections/xx18/composition-rl}{\textbf{Hugging Face Models}}
}

\begin{document}

\maketitle

\begingroup
\renewcommand{\thefootnote}{\fnsymbol{footnote}}
\footnotetext[2]{Corresponding authors.}
\endgroup

\begin{abstract}
Large-scale verifiable prompts underpin the success of Reinforcement Learning with Verifiable Rewards (RLVR), but they contain many uninformative examples and are costly to expand further.
Recent studies focus on better exploiting limited training data by prioritizing hard prompts whose rollout pass rate is 0. 
However, easy prompts with a pass rate of 1 also become increasingly prevalent as training progresses, thereby reducing the effective data size.
\revise{To mitigate this, we propose \method{}, an offline data-augmentation approach that composes existing prompts before RL to address the pass-rate-1 bottleneck without modifying the RL optimizer.}
\revise{The composed dataset is generated and verified once from the fixed source pool and then used for RL training.}
Extensive experiments across model sizes from 4B to 30B show that \method{} consistently improves reasoning capability over RL trained on the original dataset. 
Performance can be further boosted with a curriculum variant of \method{} that gradually increases compositional depth over training. 
Additionally, \method{} enables more effective cross-domain RL by composing prompts drawn from different domains.
\revise{All our code, data, and models are publicly available to facilitate future research.}
\end{abstract}

\section{Introduction}\label{sec:intro}

After the advent of OpenAI-o1~\citep{jaech2024openaio1} and DeepSeek-R1~\citep{guo2025deepseekr1}, Reinforcement Learning with Verifiable Rewards (RLVR) has reshaped the training lifecycle of large language models (LLMs), improving both text-only reasoning~\citep{deepscaler2025,yang2025qwen3,liu2025prorl,cai2025alpha-rl} and multimodal question answering~\citep{meng2025mm-eureka,xiao2025perception-r1}.
Rapid progress in RLVR, including improved optimization algorithms~\citep{liu2025prorl,nan2025ngrpo,yu2025dapo,chen2025csipo}, more efficient training frameworks~\citep{sheng2024verl,Fu2025AReaLAL,slime_github}, and techniques to mitigate training–inference mismatch~\citep{yao2025tis,qi2025fp16}, has contributed to the strong slow-thinking ability of large reasoning models (LRMs), often manifested as longer chain of thought (CoT)~\citep{CoT2022Wei}.
At its core, RLVR relies on large collections of training prompts paired with ground-truth answers to enable verifiable reward computation during training~\citep{deepscaler2025,hu2025ORZ129K,he2025deepmath}.

Prompts with $0/1$ rollout accuracy yield no gradient signals in RLVR algorithms~\citep{yu2025dapo}, substantially reducing the number of available informative training prompts. 
However, collecting and cleaning additional high-quality prompts is  expensive~\citep{he2025deepmath,zeng2025rlve}. 
To mitigate this, prior work primarily focused on hard prompts with zero success rate, via advantage shaping~\citep{nan2025ngrpo,le2025rl-zvp}, allocating more rollouts~\citep{yang2025DARS,li2025knapsackrl}, and hint-based augmentation~\citep{chen2025NuRL,li2025questa}. 
Nevertheless, while all-zero prompts constitute some fraction of the training set, as training progresses, an increasing proportion of prompts attain rollout accuracy of $1$. This motivates the need for methods that can better leverage these 'easy' prompts.

\begin{figure*}[h]
    \centering
    \includegraphics[width=0.95\linewidth]{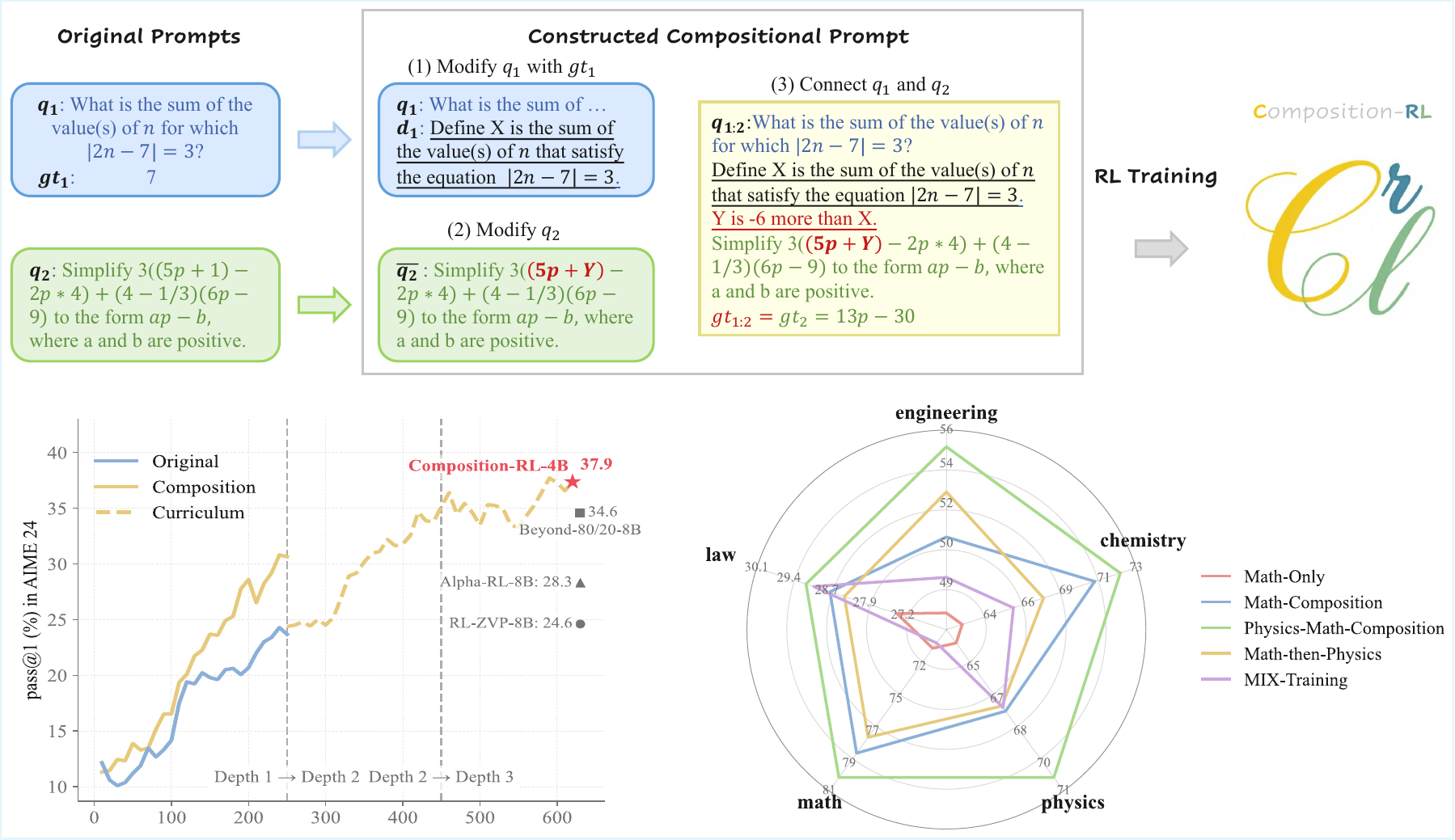}
    \caption{
    Overview of \method{}. 
\textbf{Top:} an example of composing two math problems, illustrating the high-level idea of \method{}. 
\textbf{Bottom left:} \texttt{pass@1} (\%) on AIME24 versus training steps, summarizing key findings in~\cref{subsec:finding1,subsec:finding2}. 
\textbf{Bottom right:} cross-topic results on MMLU-Pro subjects with the top-5 largest sample sizes, highlighting the main finding in~\cref{subsec:finding3}.
    }
    \label{fig:composition}
    \vspace{-0.5em}
\end{figure*}

\revise{In this work, we propose \method{}, an offline data-augmentation framework that transforms a fixed pool of existing verifiable prompts into harder training data before RL.}
\revise{Following NCSP's sequential construction and LLM-based self-verification~\citep{xiao2025ncsp}, we compose $K$ source prompts and show that composition mitigates the growing number of ``too-easy'' prompts (\cref{subsec:spc,subsec:meta-exp}).}
\revise{We then train on the fixed compositional dataset with unmodified GRPO (\cref{subsec:method_rl_composition}); an overview is provided in~\cref{fig:composition}.}
As shown in~\cref{fig:composition}, \method{} outperforms RL training on the original prompts, with increasing performance gains when combined with a curriculum over compositional depth~$K$. 
Moreover, composing prompts from different domains shows strong potential for cross-domain RL training.
Our contributions are:
\revise{\ding{182} We repurpose sequential prompt composition as offline RLVR data augmentation and introduce a pass-rate-1-motivated curriculum over compositional depth.}
\ding{183} Extensive experiments on 4B-30B LLMs demonstrate the effectiveness of \method{} and the curriculum variant of \method{}.
\ding{184} We show that RL on composed prompts spanning physics and math is more effective than simply mixing training problems, regardless of whether sequential or joint training.
\ding{185} We analyze the reasons behind the success of \method{} through the lenses of compositional generalization and implicit process supervision.

\section{Preliminary}\label{sec:preliminary}

\textbf{Notation.}
We denote an LLM parameterized by $\theta$ as a policy $\pi_\theta$.
Let $q$ be an input query (i.e., a prompt) and $\mathcal{D}$ be the set of all queries.
Given a response $r=(r_1,\ldots,r_{|r|})$ to $q$, the policy likelihood can be written as $\pi_\theta(r \mid q) = \prod_{t=1}^{|r|} \pi_\theta\!\left(r_t \mid q, r_{<t}\right)$,
where $r_{<t}=(r_1,\ldots,r_{t-1})$ and $|r|$ is the number of tokens in $r$.
Each $(q,r)$ can be evaluated by a verifier $v(q,r)\in\{0,1\}$, which indicates whether $r$ matches the ground-truth answer of $q$ (denoted as $gt$).

\textbf{RLVR.}
RLVR optimizes the expected verifiable reward:
$\max_{\theta}\, \mathbb{E}_{q \sim \mathcal{D}} \bigl[ \mathcal{J}_\text{RLVR}(\theta, q) \bigr] \, (= \mathbb{E}_{q \sim \mathcal{D},\, r \sim \pi_{\theta}(\cdot \mid q)} \bigl[v(q,r)\bigr])$.
A standard policy gradient estimator~\citep{sutton1999policygradient} is:
\begin{equation}
    g_\theta(q,r) = A(q,r) \cdot \nabla_\theta \log \pi_\theta(r|q),
    \label{eq:policy_gradient}
\end{equation}
where $A(q,r)=v(q,r)-b(q)$ is called ``advantage'' and $b(q)$ is a baseline function that depends only on the query $q$.
Group Relative Policy Optimization (GRPO)~\citep{shao2024grpo} approximates the advantage by sampling a group of $G$ responses $\{r_1,\ldots,r_G\}$ from the old policy $\pi_{\theta_{\mathrm{old}}}(\cdot\mid q)$:
\begin{equation}
\hat{A}_i
= \frac{v(q, r_i) - \mathrm{mean}\!\left(\{v(q, r_j)\}_{j=1}^{G}\right)}
{\mathrm{std}\!\left(\{v(q, r_j)\}_{j=1}^{G}\right)}.
\label{eq:advantage}
\end{equation}
Then the objective of GRPO becomes $\mathcal{J}_{\text{GRPO}}(\theta) = \mathbb{E}_{q \sim \mathcal{D}} \bigl[ \mathcal{J}_{\text{GRPO}}(\theta, q) \bigr]$ and $\mathcal{J}_{\text{GRPO}}(\theta, q)$ is:\footnote{We adopt a more-commonly used version with token-level normalization suggested by \cite{yu2025dapo}.}:
{\small
\begin{align}
\frac{1}{\sum_{i=1}^G |r_i|} \sum_{i=1}^G 
\sum_{t=1}^{|r_i|}
\min \Bigl(
    i_{i,t}(\theta)\, \hat{A}_{i},\;
    \text{clip}\bigl(i_{i,t}(\theta), 1-\epsilon, 1+\epsilon\bigr)\hat{A}_{i}
\Bigr),\, i_{i,t}(\theta)
= \frac{\pi_\theta(r_{i,t}\mid q, r_{i,<t})}{\pi_{\theta_\text{old}}(r_{i,t}\mid q, r_{i,<t})}.
\label{eq:grpo_obj} 
\end{align}
}

\begin{figure*}
    \centering
    \includegraphics[width=0.45\linewidth]{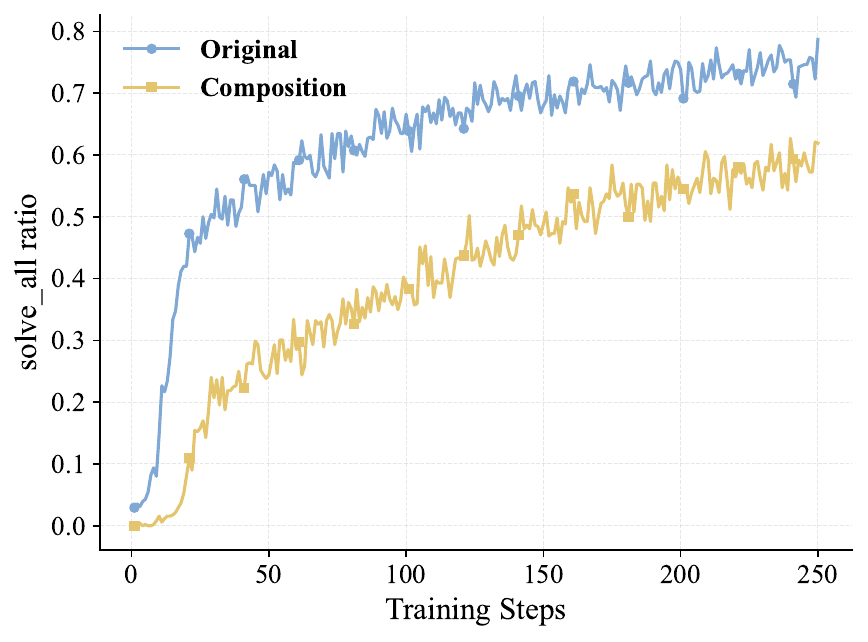}
    \includegraphics[width=0.45\linewidth]{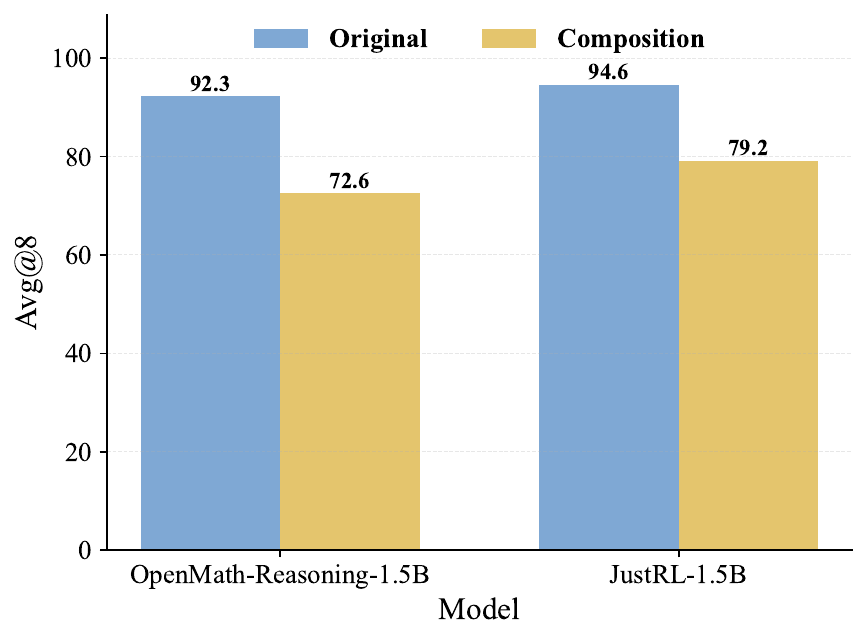}
    \caption{Visualization of meta-experiments. \textbf{Left:} \solveall{} ratio curve for RL of \texttt{Qwen3-4B-Base} with original prompts (MATH12K) versus compositional prompts. \textbf{Right:} \texttt{avg@8} accuracy on a subset of MATH500 and its corresponding compositional test prompts.}
    \label{fig:meta-exp}
    \vspace{-0.5em}
\end{figure*}

\textbf{Dynamic Sampling.}
In practice, GRPO objective can be approximated by $\hat{\mathcal{J}}_{\text{GRPO}}(\theta) = \mathbb{E}_{q \sim \mathcal{B}} \bigl[ \mathcal{J}_{\text{GRPO}}(\theta, q) \bigr]$, where $\mathcal{B} \subset \mathcal{D}$ denotes a sampled mini-batch of prompts at a given training step.
When a prompt has an empirical success rate of $0$ or $1$ (i.e., all sampled responses are incorrect or all are correct), its advantage is set to zero; consequently, by \cref{eq:policy_gradient}, policy-gradient updates vanish.
To mitigate this, dynamic sampling~\citep{yu2025dapo} first over-samples a larger candidate set $\hat{\mathcal{B}}$ and then constructs the training batch by filtering out uninformative prompts:
\begin{equation}
    \mathcal{B} = \left\{ q \in \hat{\mathcal{B}} \;:\; 0 < \mathrm{mean}\!\left(\{v(q, r_j)\}_{j=1}^{G}\right) < 1 \right\}.
    \label{eq:dynamic_sampling}
\end{equation}
Hereafter, we call a prompt \solveall{} if its sampled responses $\{r_j\}_{j=1}^{G}$ are all correct, and \solvenone{} if they are all incorrect.
Following~\citep{qu2025MoPSS,le2025rl-zvp},we use ``uninformative,'' ``zero-variance,'' and ``zero-advantage'' prompts as synonyms for \solveall{} and \solvenone{} prompts.

\section{Related Work}\label{sec:related_work}

\textbf{Longer Training with Finite Prompts.}
Amid the surge of interest in RLVR~\citep{jaech2024openaio1,guo2025deepseekr1}, many studies investigate how to enable longer and more stable training under a fixed prompt set~\citep{liu2025prorl,he2025justrl}.
One line of work improves training stability from an algorithmic perspective~\citep{chen2025csipo,yang2025entropic,liu2025deepseekv32}.
Another line aims to exploit limited training data better, including filtering uninformative prompts~\citep{yu2025dapo,qu2025MoPSS,zheng2025GRESO}, shaping advantages for zero-advantage prompts~\citep{nan2025ngrpo,le2025rl-zvp,zhu2025psrnsr}, and allocating more samples to harder prompts~\citep{yang2025DARS,li2025knapsackrl}.
Among these, hint-based problem augmentation~\citep{chen2025NuRL,li2025questa} is most closely related to our work: they use hints to transform originally hard prompts into easier ones.
In contrast, we make easy prompts harder via compositional prompt generation.

\textbf{Enlarging RLVR Training Prompts.}
The fuel of RLVR is its training prompts.
A substantial body of work is devoted to collecting and curating high-quality data from diverse sources~\citep{hu2025ORZ129K,he2025deepmath,albalak2025bigmath}.
\revise{Synthesizing data from existing datasets has also been extensively studied, both for evaluation~\citep{shi2023gsm-ic,xu2024egsm} and for SFT~\citep{metamath2023yu,yang2024qwen25math,dartmath2024tong}.}
\revise{NCSP~\citep{xiao2025ncsp} introduces the sequential construction and LLM-based self-verification used for mathematical evaluation. 
We follow that construction but use its pipeline to augment RL data, adding a pass-rate-1-motivated curriculum, cross-domain composition, and analyses of the resulting gains.}
\revise{MathFusion~\citep{pei2025mathfusion} fuses multiple instructions for SFT, whereas MathForge~\citep{dai2026mathforge} combines a difficulty-aware RL objective with single-question reformulation. In contrast, we retain the standard RL optimizer and compose multiple source prompts to reuse otherwise uninformative \solveall{} examples.}
More recently, several efforts have begun to synthesize prompts specifically for RL training~\citep{zeng2025rlve,xie2025logicrl,li2025internbootcamp,stojanovski2025reasoninggym}.
\revise{Unlike synthetic logic-only problems or game-like environments, our experiments focus on mathematical RL and cross-domain composition.}

\textbf{Compositional Generalization.}
Compositional generalization refers to a model’s ability to recombine learned skills to solve novel tasks.
It has been a longstanding focus in natural language processing~\citep{keysers2019composition1,hupkes2020composition2,lake2018compostion3}.
Prior work often studies compositionality using controlled testbeds, such as Skill-Mix~\citep{yu2023skillmix} for language tasks, compositional math benchmarks~\citep{sun2025omega}, or algorithmic tasks~\citep{dziri2023compositiontoy1}.
\citep{zhao2024compositionsft} show that composing textual skills can benefit SFT.
\citep{yuan2025rl-composition} suggest that compositionality is important for RL to acquire new skills.
However, their results are restricted to synthesized string-manipulation tasks.
\revise{In comparison, we study compositional prompts as RL training data across model scales and analyze compositional generalization and implicit process supervision.}

\section{Methodology \& Meta-Experiments}\label{sec:method}

\subsection{\datamethod{}: Sequential Prompt Composition}\label{subsec:spc}

\citep{yuan2025rl-composition} studies the role of composition in RL using a synthetic string-transformation setting, and \citep{xiao2025ncsp} evaluates LLM performance under the composition of two math problems. We extend this line of work by investigating how composing training prompts affects RL training. In this section, we describe Sequential Prompt Composition (\datamethod{}): we first define how to compose two prompts, and then generalize to composing $K$ prompts.
The whole composition process is illustrated in \cref{fig:composition}.

\textbf{Composing Two Prompts.}
Given two prompts $q_1$ and $q_2$ with ground-truth answers $gt_1$ and $gt_2$, we define a composition operator $\compose{}$ that maps $(q_1,q_2; gt_1,gt_2)$ to a composed prompt $q_{1:2}$ with ground-truth answer $gt_{1:2}$:
\begin{equation}
   q_{1:2}, \, gt_{1:2} = \compose{} (q_1, \,q_2;\, gt_1,\, gt_2).
\end{equation}
\revise{This operator only requires $q_1$ to have an extractable numeric answer and $q_2$ to contain a replaceable numeric condition; neither source prompt must itself be a mathematics problem.}
The operator $\compose{}$ consists of three steps (see~\cref{fig:composition}(a) for one concrete example):

\ding{182} \textbf{Modify $q_1$ with $gt_1$.}
Extract a numeric value from $gt_1$, denoted by $v_1$. We then introduce a natural-language definition $d_1$ that names this value in terms of $(q_1,gt_1)$, and form $\bar{q}_1 = q_1 \oplus d_1$.
For instance, if
$q_1$ is ``\textit{What is the sum of the value(s) of $n$ for which $|2n-7|=3$?}'' and $gt_1=7$,
we set $v_1=7$ and add a definition such as:
``\textit{Let $X$ be the sum of the value(s) of $n$ satisfying $|2n-7|=3$.}''

\ding{183} \textbf{Modify $q_2$.}
Extract a numeric value from $q_2$ and replace it with a new variable $v_2$, yielding $\bar{q}_2 = q_2(v_2)$.
For example, if
$q_2$ is ``\textit{Simplify $2((5p+1)-2p\cdot 4)+(4-1\div 3)(6p-9)$ to the form $ap-b$, where $a$ and $b$ are positive,}''
we may choose the constant $1$ as $v_2$ and replace it with variable name $Y$, obtaining
``\textit{Simplify $2((5p+{\color{red}{Y}})-2p\cdot 4)+(4-1\div 3)(6p-9)$ to the form $ap-b$...}''

\ding{184} \textbf{Connect $q_1$ and $q_2$.}
Compute $v_1 - v_2$ and express the relation between the two variables as a statement $r$. 
Continuing the example above, with $v_1=7$ and $v_2=1$, we have $v_1-v_2=6$, so we can add a constraint $r$ such as:
``\textit{$Y$ is $6$ less than $X$.}'' 
The composed prompt is then $q_{1:2} = \bar{q}_1 \oplus r \oplus \bar{q}_2$.
By construction, the answer of the composed prompt is $gt_{1:2}=gt_2$. 
This composition is asymmetric to the order of $q_1$ and $q_2$, and solving $q_{1:2}$ requires solving $q_1$ first and then $q_2$.

\textbf{Composing $K$ Prompts.}
More generally, we can compose $K$ prompts into one prompt. Given $q_1,\dots,q_K$ with ground-truth answers $gt_1,\dots,gt_K$, Sequential Prompt Composition (\datamethod{}) applies $\compose{}$ recursively for $K-1$ steps:
{\small
\begin{align*}
&\datamethod{}(q_1,\dots,q_K;\, gt_1,\dots,gt_K)
= \compose{}(q_1,\, q_{2:K};\, gt_1,\, gt_{2:K}), \\
& \text{where} \quad (q_{2:K}, gt_{2:K})
= \datamethod{}(q_2,\dots,q_K;\, gt_2,\dots,gt_K).
\end{align*}
}Finally, we will get the composed prompt $q_{1:K}$ and its answer $gt_{1:K}$.
We term $K$ as the \textit{Compositional Depth}.
Intuitively, solving $q_{1:K}$ requires the model having the ability to solve all $\{q_k\}_{k=1}^K$.
The process of composing $2$ prompts can be viewed as a special case of \datamethod{} with $K=2$.
Therefore, we do not distinguish between these two hereinafter.

\subsection{Meta-Experiments \& Observation}\label{subsec:meta-exp}

As collecting new high-quality, verifiable training prompts can be costly~\citep{he2025deepmath,zeng2025rlve}, a growing body of work has focused on better leveraging uninformative prompts in RLVR~\citep{li2025questa,li2025knapsackrl}. 
However, existing methods primarily target \solvenone{} prompts. 
In this section, we conduct some meta-experiments and have the following key observations: \ding{182} Beyond \solvenone{}, the increasing prevalence of \solveall{} prompts is another major impediment to effective RL training.
\ding{183} \datamethod{} can make easy prompts harder and reduce the ratio of \solveall{} prompts.

\textbf{Dilemma of Effective Training Prompts.}
As the policy model becomes stronger during RLVR, the proportion of \solveall{} prompts observed during rollouts increases. 
\cref{fig:meta-exp} (Left) plots the \solveall{} rate across RL training steps for \texttt{Qwen3-4B-Base} on the MATH training set~\citep{hendrycks2021math}. 
The \solveall{} ratio rises rapidly from near zero to over 50\% within the first 50 steps and then stabilizes around 75\%. 
Although dynamic sampling is enabled to remove zero-variance prompts, the actual effective size of the whole training set at later stages is reduced to roughly 3K prompts ($12{,}000 \times (1 - 0.75)$). 
In contrast, the \solvenone{} ratio remains low (about 5\%) at 250 steps. 
These results motivate methods that can deal with \solveall{} prompts, in addition to \solvenone{} prompts.

\textbf{\datamethod{} can nudge the last bits out of existing prompts.}
Intuitively, \datamethod{} makes easy prompts harder. 
We empirically validate this on a subset of the MATH500 test set using \texttt{OpenMath-Reasoning-1.5B}~\citep{moshkov2025openmathreasoning} and \texttt{JustRL-1.5B}~\citep{he2025justrl}; additional details are provided in \cref{subapp:meta-exp2}. 
As shown in \cref{fig:meta-exp} (Right), switching to compositional prompts reduces $avg@8$ by $19.7\%$ for \texttt{OpenMath-Reasoning-1.5B} and by $15.4\%$ for \texttt{JustRL-1.5B}. 
The \solveall{} rate also drops substantially: from $81.5\%$ to $41.4\%$ for \texttt{OpenMath-Reasoning-1.5B}, and from $88.5\%$ to $60.0\%$ for \texttt{JustRL-1.5B}. 
These results suggest that \datamethod{} can effectively reduce \solveall{} prompts, potentially turn part of the original uninformative prompts useful again.
\revise{Additionally, at $K=2$, the ordered candidate space contains up to $|\mathcal{D}|\cdot(|\mathcal{D}|-1)$ compositions and thus grows quadratically with the number of source prompts.}

Note that \texttt{JustRL-1.5B} is obtained by RL training \texttt{OpenMath-Reasoning-1.5B}. 
Another interesting observation is that \texttt{JustRL-1.5B} improves performance both on the MATH500 subset (by 2.3\%) and the compositional test set (by 6.6\%). 
This suggests that RL training on normal prompts\footnote{As compositional prompts differ substantially in structure from the original ones, we refer to prompts from all existing datasets as \emph{normal} prompts for simplicity.} can also improve performance on compositional prompts.
This raises a natural question: \textit{Does RL training on compositional prompts benefit performance on normal reasoning tasks?}

\subsection{\method{}: RL with Compositional Data}\label{subsec:method_rl_composition}

\revise{\method{} is an offline data-augmentation framework rather than a new RL optimizer.}
\revise{The growing \solveall{} rate motivates composition and the depth curriculum, but does not define an online selection rule: every prompt in the fixed source pool remains eligible for composition.}
Given the original training set $\mathcal{D}=\{(q_i, gt_i)\}_{i=1}^{|\mathcal{D}|}$, we can construct a level-$K$ compositional prompt set via the LLM-driven \datamethod{} procedure:
{\small
\begin{align*}
\mathcal{D}_{C_K}
   = \bigl\{(q, gt)\,:\, q,gt=\datamethod{}(q_1,\dots,q_K;\, gt_1,\dots,gt_K),\; 
  (q_k,gt_k)\in\mathcal{D},\,\;
  q_i\neq q_j~\forall i\neq j \bigr\}.
\end{align*}
}Since the size of $\mathcal{D}_{C_K}$ can be extremely large, we instead use a smaller surrogate set:
{\small
\begin{align}
\hat{\mathcal{D}}_{C_K}
  = \bigl\{(q, gt)\,:\, q,gt=\datamethod{}(q_1,\dots,q_K;\, gt_1,\dots,gt_K),\;
  (q_k,gt_k)\in\mathcal{D}_k,\, \;
  q_i\neq q_j~\forall i\neq j \bigr\},
  \label{eq:composition_data}
\end{align}
}where each $\mathcal{D}_k$ is a small random subset of $\mathcal{D}$ and serve as the candidate set for $q_k$, i.e., $q_k \sim \mathcal{D}_k$.
In practice, we set $|\mathcal{D}_k|=20$ for $k=1,\dots,K-1$, and $\mathcal{D}_K=\mathcal{D}$.
\revise{During this offline construction, each new $q_K\in\mathcal{D}$ is paired with independently resampled candidate subsets $\mathcal{D}_1,\dots,\mathcal{D}_{K-1}$; after generation and verification, $\hat{\mathcal{D}}_{C_K}$ is fixed throughout RL training.}

\method{} optimizes the objective:
\revise{$\max_{\theta}\; \mathbb{E}_{q \sim \hat{\mathcal{D}}_{C_K}}\bigl[\mathcal{J}_{\text{RLVR}}(\theta,q)\bigr]$.}
\revise{We use the GRPO objective $\mathcal{J}_{\text{GRPO}}(\theta)$, advantage estimator, and importance ratio in \cref{eq:grpo_obj,eq:advantage}; only the prompt distribution changes to the fixed compositional dataset.}
\revise{Dynamic sampling during RL merely filters zero-variance prompts from training batches and does not generate new compositions online.}
Unless otherwise specified, we use $K=2$ in all remaining experiments, and abbreviate $\mathcal{D}_{C_2}$ as $\mathcal{D}_C$.

\section{Experiments}\label{sec:exp}

\begin{table*}[!t]
    \centering
    \caption{%
    Main Results.
``\texttt{Avg@k}'' denotes the average accuracy (\%) over $k$ random generations (i.e., \texttt{pass@1}).
The rows ``Depth 1 + 2'' and ``+ Depth 3'' are the results of curriculum \method{} in~\cref{subsec:finding2}.
\revise{$^*$ indicates that the corresponding results are obtained using the same training set size as MATH12K.
For {\method}, the starred row uses a 12K subset of {\dataset}, while the unstarred row uses the full set.}
}
\vspace{-1em}
    \label{tab:main_result}
    \vspace{2mm}
    \resizebox{0.95\textwidth}{!}{
\begin{tabular}{c|ccccc|ccc|c}
\toprule[1.6pt]
    \multirow{3}{*}{\revise{\textbf{Method / Data}}} 
    & \multicolumn{5}{c|}{\textbf{Mathematics (In-Domain)}} 
    & \multicolumn{3}{c|}{\textbf{Multi-Task (Out-Of-Domain)}}
    & \textbf{Overall} \\ \cmidrule(lr){2-10}
    ~ & \textbf{AIME 24} 
      & \textbf{AIME 25} 
      & \textbf{Beyond AIME} 
      & \textbf{IMOBench}
      & \textbf{Overall}
      & \textbf{GPQA}
      & \textbf{MMLU-Pro}
      & \textbf{Overall}
      & \textbf{Overall} \\ 
    & \texttt{Avg@32} & \texttt{Avg@32} & \texttt{Avg@8} & \texttt{Avg@4}
      & \texttt{Avg.}
      & \texttt{Avg@8} & \texttt{Avg@1}
      & \texttt{Avg.}
      & \texttt{Avg.} \\ 
\midrule
\multicolumn{10}{c}{\texttt{\textbf{{Qwen3-4B-Base}}}} \\
\midrule
 \revise{GRPO} & 23.3 & 19.5 & 9.0 & 14.4 & 16.6 & 43.7 & 58.6 & 51.2 & 28.1 \\
 MetaMath$^*$ & 12.7 & 9.7 & 3.9 & 8.1 & 8.6 & 43.9 & 58.5 & 51.2 &22.8 \\
 SAND-Math$^*$ & 25.6 & 20.5 & 8.9 & 15.4 & 17.6 & 45.4 & 57.9 & 51.7 & 29.0 \\
 \revise{MathFusion$^*$} & \revise{16.5} & \revise{9.5} & \revise{6.3} & \revise{8.9} & \revise{10.3} & \revise{41.5} & \revise{54.2} & \revise{47.9} & \revise{22.8} \\
 \revise{MathForge$^*$} & \revise{25.5} & \revise{14.1} & \revise{6.0} & \revise{10.9} & \revise{14.1} & \revise{36.8} & \revise{47.4} & \revise{42.1} & \revise{23.4} \\
 \rowcolor[rgb]{ .867, .922, .969} \revise{\method{}$^*$} & 27.3\up{4.0} & 23.1\up{3.6} & 9.0\up{0.0} & 15.9\up{1.5} & \revise{18.8\up{2.2}} & 46.2\up{2.5} & 60.1\up{1.5} & 53.2\up{2.0} & 30.3\up{2.2} \\
\rowcolor[rgb]{ .867, .922, .969} \revise{\method{}} & 30.5\up{7.2} & 23.3\up{3.8} & 12.6\up{3.6} & 14.3\down{0.1} & 20.2\up{3.6} & 46.3\up{2.6} & 61.4\up{2.8} & 53.9\up{2.7} & 31.4\up{3.3} \\
\midrule
\rowcolor[rgb]{ .867, .922, .969} Depth 1 + 2 & 33.0\up{9.7} & 27.8\up{8.3} & 13.1\up{4.1} & 20.1\up{5.7} & 23.5\up{6.9} & 48.3\up{4.6} & 63.8\up{5.2} & 56.1\up{4.9} & 34.4\up{6.3} \\
\rowcolor[rgb]{ .867, .922, .969} + Depth 3 & 37.9\up{14.6} & 29.7\up{10.2} & 14.6\up{5.6} & 22.9\up{8.5} & 26.3\up{9.7} & 48.5\up{4.8} & 64.5\up{5.9} & 56.5\up{5.3} & 36.4\up{8.3} \\
\midrule
\multicolumn{10}{c}{\texttt{\textbf{{Qwen3-8B-Base}}}} \\
\midrule
\revise{GRPO} & 26.1 & 20.4 & 13.7 & 16.2 & 19.1 & 48.2 & 62.6 & 55.4 & 31.2 \\
\rowcolor[rgb]{ .867, .922, .969} \revise{\method{}$^*$} & 29.2\up{3.1} & 25.2\up{4.8} & 13.3\down{0.4} & 18.8\up{2.6} & 21.6\up{2.5} & 49.1\up{0.9} & 65.9\up{3.3} & 57.5\up{2.1} & 33.6\up{2.4} \\
\rowcolor[rgb]{ .867, .922, .969} \revise{\method{}} & 36.9\up{10.8} & 26.5\up{6.1} & 13.9\up{0.2} & 18.4\up{2.2} & 23.9\up{4.8} & 48.9\up{0.7} & 64.5\up{1.9} & 56.7\up{1.3} & 34.9\up{3.7} \\
\midrule
\multicolumn{10}{c}{\texttt{\textbf{{Qwen3-14B-Base}}}} \\
\midrule
 \revise{GRPO} & 34.4 & 30.2 & 17.0 & 21.3 & 25.7 & 55.0 & 67.2 & 61.1 & 37.5 \\
 \rowcolor[rgb]{ .867, .922, .969} \revise{\method{}$^*$} & 43.3\up{8.9} & 36.0\up{5.8} & 18.6\up{1.6} & 23.5\up{2.2} & 30.4\up{4.7} & 54.7\down{0.3} & 69.3\up{2.1} & 62.0\up{0.9} & 40.9\up{3.4} \\
\rowcolor[rgb]{ .867, .922, .969} \revise{\method{}} & 44.5\up{10.1} & 36.9\up{6.7} & 19.7\up{2.7} & 25.9\up{4.6} & 31.8\up{6.1} & 54.2\down{0.8} & 69.3\up{2.1} & 61.8\up{0.7} & 41.8\up{4.3} \\
\midrule
\multicolumn{10}{c}{\texttt{\textbf{{Qwen3-30B-A3B-Base}}}} \\
\midrule
 \revise{GRPO} & 25.2 & 16.2 & 7.5 & 13.2 & 15.5 & 50.7 & 62.6 & 56.7 & 29.2 \\
 \rowcolor[rgb]{ .867, .922, .969} \revise{\method{}$^*$} & 47.7\up{22.5} & 29.8\up{13.6} & 20.1\up{12.6} & 22.9\up{9.7} & 30.1\up{14.6} & 58.2\up{7.5} & 65.6\up{3.0} & 61.9\up{5.2} & 40.7\up{11.5} \\
\rowcolor[rgb]{ .867, .922, .969} \revise{\method{}} & \revise{46.4\up{21.2}} & 30.3\up{14.1} & 19.5\up{12.0} & 22.8\up{9.6} & 29.8\up{14.3} & 54.6\up{3.9} & 64.6\up{2.0} & 59.6\up{2.9} & 39.7\up{10.5} \\
\bottomrule[1.6pt]
\end{tabular}}
\vspace{-0.5em}
\end{table*}

\subsection{Experimental Setting}\label{subsec:exp_setup}

In this section, we briefly summarize our experimental setup, including training procedures, baselines, and evaluation. Additional details are provided in \cref{app:exp_detail}.

\textbf{Training Details}
We conduct RL training using the VeRL codebase~\citep{sheng2024verl}. Unless otherwise specified, we use a unified set of hyperparameters (batch size $256$, learning rate $1\times 10^{-6}$, and no warm-up) and fixed rollout settings (temperature $1$, \texttt{top\_p} $1$, \texttt{top\_k} $-1$, 8 rollouts per problem, and a maximum output length of 16K tokens). 
We train \texttt{Qwen3-4B-Base}, \texttt{Qwen3-8B-Base}, \texttt{Qwen3-14B-Base}, and \texttt{Qwen3-30B-A3B-Base} on the MATH training set~\citep{hendrycks2021math}.
For the cross-topic experiments in \cref{subsec:finding3}, we use the physics subset of MegaScience~\citep{fan2025megascience}.
For the verifier, we choose \href{https://github.com/huggingface/Math-Verify}{Math-Verify}, a rule-based verifier.
For a fair comparison, we enable dynamic sampling to filter uninformative prompts, ensuring that the effective batch size at each step remains constant across experiments.

\textbf{Baselines.}
In \cref{subsec:finding1} (see \cref{tab:main_result}), we compare \method{} with standard RLVR on MATH12K under the same number of gradient updates. For \method{}, we construct approximately 199K compositional prompts, which we denote as \dataset{}.
\revise{
For the 4B setting, we also compare with data-augmentation baselines, including MetaMath~\citep{metamath2023yu,lu2024mathgenie}, SAND-Math~\citep{manem2025sandmath}, MathFusion~\citep{pei2025mathfusion}, and MathForge~\citep{dai2026mathforge}.
For fairness, we additionally report results of {\method} under a controlled training set size (denoted by ``*'' in~\cref{tab:main_result}).
}
For the cross-domain experiments in \cref{subsec:finding3}, we compare \method{} with two baselines: \textit{Mix Training} (RL on a mixed dataset comprising MATH12K and the MegaScience Physics subset) and \textit{Math-then-Physics} (continued RL on Physics starting from a MATH12K-trained checkpoint). 
Additional details are provided in \cref{subapp:baseline}.

\textbf{Evaluation Details.}
Our evaluation benchmarks include both in-domain (ID) math reasoning tasks, AIME24/25, BeyondAIME~\citep{bytedance_seed_2025_beyondaime}, and IMOBench~\citep{luong2025imobench}, and out-of-domain (OOD) multi-task reasoning benchmarks, GPQA-Diamond~\citep{rein2024gpqa} and MMLU-Pro~\citep{wang2024mmlupro}. Following \citep{guo2025deepseekr1}, we sample multiple responses per problem (from 1 to 32, depending on the benchmark size) and report \texttt{pass@1} accuracy. 
Following~\citep{yang2025qwen3,deepscaler2025,xu2025thinking}, we set the temperature to 0.6, \texttt{top\_p} to 0.95, \texttt{top\_k} to 20, and the maximum output length to 32K tokens.
See also \cref{subapp:evaluation} for details.

\subsection{Compositional Prompts Are Beneficial to RLVR}\label{subsec:finding1}

To evaluate \method{}, we report main results in~\cref{tab:main_result} \revise{with additional metrics provided in~\cref{app:more_results}.}
From~\cref{tab:main_result}, we have the following observations:

\revise{\ding{182} \textbf{RL on compositional prompts improves aggregate in-domain mathematics and OOD multi-task performance across all model sizes.}}  
Across all model sizes, \method{} improves the overall mathematics performance by $+3.6\%$, $+4.8\%$, $+6.1\%$, and $+14.3\%$ for \texttt{Qwen3-4B/8B/14B/30B-A3B}, respectively. 
\revise{With the full compositional set, gains reach $+21.2\%$ on AIME24, $+14.1\%$ on AIME25, $+12.0\%$ on Beyond AIME, and $+9.6\%$ on IMOBench.}
Moreover, \method{} also improves OOD performance, increasing the multi-task overall by $+2.7\%$, $+1.3\%$, $+0.7\%$, and $+2.9\%$, leading to overall average gains of $+3.3\%$, $+3.7\%$, $+4.3\%$, and $+10.5\%$ across the four base models.
\revise{These substantial aggregate gains demonstrate the effectiveness of \method{}.}%

\ding{183} \textbf{The benefits of \method{} scale with model size, with larger models exhibiting substantially larger improvements, especially in mathematics.}
Overall gains increase from $+3.3\%$ (4B) and $+3.7\%$ (8B) to $+4.3\%$ (14B), and peak at $+10.5\%$ for \texttt{Qwen3-30B-A3B}. 
The scaling effect is most pronounced on in-domain mathematics: improvements rise from $+3.6\%$/$+4.8\%$/$+6.1\%$ to $+14.3\%$ as model size increases from 4B to 30B, whereas OOD multi-task gains are smaller but remain consistently positive.  
Notably, the MoE \texttt{30B-A3B} model underperforms the 14B dense model, consistent with the fact that MoE activates only a subset of experts per token and can be more sensitive to routing and optimization under a fixed training budget; 
nevertheless, \method{} still yields large gains on this model. 
Overall, these results highlight the scaling potential of \method{}.
\revise{\ding{184} \textbf{{\method} still outperforms all baselines under the same training set size.}
Compared with RL training on the original dataset, {\method} with 12K compositional prompts yields gains ranging from $2.2\%$ to $11.5\%$ for models from 4B to 30B.
For the 4B model, {\method} outperforms MetaMath, SAND-Math, MathFusion, and MathForge; the closest baseline, SAND-Math, is lower by $1.3\%$ overall.
This indicates that the gains are not explained solely by increasing the number of training prompts, even though the seed questions are drawn from the same MATH training set.
Overall, {\method} achieves the best performance under a controlled dataset size, and its ability to generate substantially more training prompts from existing human-annotated data is another key advantage.}

\begin{table*}[!t]
    \centering
    \caption{Results of cross-topic experiments across multiple benchmarks. ``\texttt{Avg@k}'' denotes the average accuracy (\%) over $k$ random generations (i.e., \texttt{pass@1}). ``\texttt{MATH12K} + \texttt{Physics}'' corresponds to the \textit{Mix Training} baseline, and ``\texttt{Physics} after \texttt{MATH12K}'' corresponds to the \textit{Math-then-Physics} baseline. \revise{The final block uses 10K multidisciplinary prompts from SCP-116K~\citep{lu2025scp116k}.} %
}
\vspace{-1em}
    \label{tab:cross_topic}
    \vspace{2mm}
    \resizebox{0.95\textwidth}{!}{
\begin{tabular}{c|ccccc|ccc|c}
\toprule[1.6pt]
    \multirow{3}{*}{\textbf{Dataset}} 
    & \multicolumn{5}{c|}{\textbf{Mathematics}} 
    & \multicolumn{3}{c|}{\textbf{Multi-Task}}
    & \textbf{Overall} \\ \cmidrule(lr){2-10}
    ~ & \textbf{AIME 24} 
      & \textbf{AIME 25} 
      & \textbf{Beyond AIME} 
      & \textbf{IMOBench}
      & \textbf{Overall}
      & \textbf{GPQA}
      & \textbf{MMLU-Pro}
      & \textbf{Overall}
      & \textbf{Overall} \\ 
    & \texttt{Avg@32} & \texttt{Avg@32} & \texttt{Avg@8} & \texttt{Avg@4}
      & \texttt{Avg.}
      & \texttt{Avg@8} & \texttt{Avg@1}
      & \texttt{Avg.}
      & \texttt{Avg.} \\ 
      \midrule
 \texttt{MATH12K} & 23.3 & 19.5 & 9.0 & 14.4 & 16.6 & 43.7 & 58.6 & 51.2 & 28.1 \\
\midrule
\texttt{MATH12K} + \texttt{Physics} & 19.7 & 16.5 & 8.3 & 12.0 & 14.1 & 44.4 & 59.6 & 52.0 & 26.8 \\
\texttt{Physics} after \texttt{MATH12K} & 25.3 & 22.3 & 8.6 & 14.4 & 17.7 & 45.2 & 61.4 & 53.3 & 29.5 \\
\rowcolor[rgb]{ .867, .922, .969} \datasetphysicsmath{} & \textbf{32.4} & \textbf{25.5} & \textbf{10.6} & \textbf{17.8} & \textbf{21.6} & \textbf{46.6} & \textbf{62.7} & \textbf{54.7} & \textbf{32.6} \\
\midrule
\multicolumn{10}{c}{\revise{\texttt{SCP-116K Multidisciplinary Prompts (10K)}}} \\
\midrule
\revise{Mix RL} & \revise{10.10} & \revise{9.79} & \revise{4.25} & \revise{8.31} & \revise{8.11} & \revise{44.87} & \revise{60.58} & \revise{52.73} & \revise{22.98} \\
\rowcolor[rgb]{ .867, .922, .969} \revise{\method{}} & \revise{13.02} & \revise{11.10} & \revise{5.38} & \revise{9.69} & \revise{9.80} & \revise{49.84} & \revise{65.00} & \revise{57.42} & \revise{25.67} \\

\bottomrule[1.6pt]
\end{tabular}}
\vspace{-0.5em}
\end{table*}

\begin{table*}[t]
    \centering
    \caption{%
    \revise{Results of the ablation study.}
$\mathcal{D}_1$ specifies the strategy for constructing the candidate set for $q_1$: ``RAND'' randomly samples 20 prompts, whereas ``FULL'' selects from the entire original prompt set $\mathcal{D}$.  
``Baseline'' denotes RL training on the original prompts $\mathcal{D}$.
\revise{``Direct Concat'' denotes RL training on prompts that are directly concatenated from two prompts without composition.
For results on the fully solvable subset, n is the number of rollouts.
Additional blocks report the Feeding-$v_1$ control, standalone training at each composition depth, and adaptive Depth-2 composition.}}
\vspace{-1em}
    \label{tab:ablation}
    \vspace{2mm}
    \resizebox{0.95\textwidth}{!}{
\begin{tabular}{cc|ccccc|ccc|c}
\toprule[1.6pt]
    \multirow{3}{*}{\textbf{$\mathcal{D}_1$}} & \multirow{3}{*}{\textbf{$\mathcal{D}_2$}} 
    & \multicolumn{5}{c|}{\textbf{Mathematics (In-Domain)}} 
    & \multicolumn{3}{c|}{\textbf{Multi-Task (Out-Of-Domain)}}
    & \textbf{Overall} \\ \cmidrule(lr){3-11}
    ~& & \textbf{AIME 24} 
      & \textbf{AIME 25} 
      & \textbf{Beyond AIME} 
      & \textbf{IMOBench}
      & \textbf{Overall}
      & \textbf{GPQA}
      & \textbf{MMLU-Pro}
      & \textbf{Overall}
      & \textbf{Overall} \\ 
   & & \texttt{Avg@32} & \texttt{Avg@32} & \texttt{Avg@8} & \texttt{Avg@4}
      & \texttt{Avg.}
      & \texttt{Avg@8} & \texttt{Avg@1}
      & \texttt{Avg.}
      & \texttt{Avg.} \\ 
      \midrule
\multicolumn{2}{c|}{Baseline} & 23.3 & 19.5 & 9.0 & \textbf{14.4} & 16.6 & 43.7 & 58.6 & 51.2 & 28.1 \\
RAND & RAND & 22.6 & 19.6 & 8.2 & 13.8 & 16.1 & 43.6 & 60.0 & 51.8 & 28.0 \\
FULL & RAND & 24.5 & \textbf{23.4} & 8.7 & 14.1 & 17.7 & 44.8 & 59.9 & 52.4 & 29.2 \\
\rowcolor[rgb]{ .867, .922, .969}  RAND & FULL & \textbf{30.5} & 23.3 & \textbf{12.6} & 14.3 & \textbf{20.2} & \textbf{46.3} & \textbf{61.4} & \textbf{53.9} & \textbf{31.4} \\
\midrule
\multicolumn{2}{c|}{Direct Concat 12K} & 18.4 & 14.4 & 6.4 & 10.4 & 12.4 & 43.1 & 55.9 & 49.5 & 24.8 \\
\rowcolor[rgb]{ .867, .922, .969} \multicolumn{2}{c|}{\method{} 12K} & \textbf{27.3} & \textbf{23.1} & \textbf{9.0} & \textbf{15.9} & \textbf{18.8} & \textbf{46.2} & \textbf{60.1} & \textbf{53.2} & \textbf{30.3} \\
\midrule
\multicolumn{2}{c|}{\revise{Feeding-$v_1$}} & \revise{22.8} & \revise{19.8} & \revise{9.2} & \revise{13.9} & \revise{16.4} & \revise{44.0} & \revise{58.1} & \revise{51.1} & \revise{28.0} \\
\midrule
\multicolumn{11}{c}{\revise{\texttt{\textbf{Standalone Composition Depth and Adaptive Composition}}}} \\
\midrule
\multicolumn{2}{c|}{\revise{Depth 1}} & \revise{23.3} & \revise{19.5} & \revise{9.0} & \revise{14.4} & \revise{16.6} & \revise{43.7} & \revise{58.6} & \revise{51.2} & \revise{28.1} \\
\multicolumn{2}{c|}{\revise{Depth 2}} & \revise{27.3} & \revise{23.1} & \revise{9.0} & \revise{15.9} & \revise{18.8} & \revise{46.2} & \revise{60.1} & \revise{53.2} & \revise{30.3} \\
\multicolumn{2}{c|}{\revise{Depth 3}} & \revise{26.1} & \revise{22.3} & \revise{9.2} & \revise{15.2} & \revise{18.2} & \revise{45.9} & \revise{60.2} & \revise{53.1} & \revise{29.8} \\
\rowcolor[rgb]{ .867, .922, .969} \multicolumn{2}{c|}{\revise{Adaptive Depth 2}} & \revise{28.2} & \revise{22.3} & \revise{9.2} & \revise{16.3} & \revise{19.0} & \revise{46.7} & \revise{59.9} & \revise{53.3} & \revise{30.4} \\
\midrule
\multicolumn{11}{c}{\texttt{\textbf{{Results on Fully Solvable Subset of MATH12K}}}} \\
\midrule
\multicolumn{2}{c|}{initial model} & 13.1 & 10.1 & 5.2 & 8.7 & 9.3 & 42 & 53.9 & 48.0 & 22.2 \\
\multicolumn{2}{c|}{Baseline (n=32)} & 13.2 & 11.6 & 5.1 & 9.9 & 10.0 & 42.1 & 53.7 & 47.9 & 22.6 \\
\rowcolor[rgb]{ .867, .922, .969}\multicolumn{2}{c|}{{\method} (n=8)} & \textbf{17.1} & \textbf{15.6} & \textbf{6.1} & \textbf{10.5} & \textbf{12.3} & \textbf{42.5} & \textbf{57.6} & \textbf{50.1} & \textbf{24.9} \\
\bottomrule[1.6pt]
\end{tabular}}
\vspace{-1em}
\end{table*}

\subsection{Curriculum RL to Higher Compositional Depth}\label{subsec:finding2}

We experiment with a curriculum version of \method{} that progressively increases the composition depth and continues RL training.
As shown in~\cref{tab:main_result} and~\cref{fig:composition}, we have:

\ding{182} \textbf{Curriculum \method{} can make full use of the original prompts, producing progressively stronger LRMs as the composition depth increases.}
\revise{Continuing RL with Depth~2 data after Depth~1 (i.e., the original MATH12K) yields substantial gains over the Depth~1 checkpoint, improving by $+9.7\%$ on AIME24 and $+5.2\%$ on MMLU-Pro.}
Moreover, the Depth~1$\rightarrow$Depth~2 curriculum even outperforms training directly on \dataset{}, delivering a further $+3.0\%$ improvement on average. 
Adding Depth~3 stage continues to improve both in-domain tasks and OOD questions, with a further $+2.0\%$ overall gain. 
\Cref{fig:composition} presents the validation performance curves throughout the curriculum training process. 
In summary, these results imply that \method{} effectively converts limited prompts (with high \solveall{} rates) into more useful samples.

\ding{183} \revise{\textbf{Standalone depth and adaptive composition.}
The standalone controls in~\cref{tab:ablation} improve from $28.1\%$ overall at Depth~1 to $30.3\%$ at Depth~2, while Depth~3 obtains $29.8\%$; thus, increasing depth alone does not yield monotonic gains.
In the adaptive variant, when current-policy rollouts identify a source prompt as \solveall{}, we replace it with a pre-generated composition whose second constituent $q_2$ is that source prompt.
Adaptive Depth~2 composition reaches $30.4\%$, essentially matching the standard Depth~2 result of $30.3\%$, so this simple adaptive construction provides no clear advantage over the offline construction.}

\ding{184} \textbf{Curriculum \method{} on a 4B model surpasses several 8B baselines, even under unfavorable settings.}
As shown in~\cref{fig:composition}, our final \texttt{Composition-RL-4B} model achieves $37.9\%$ on AIME24, outperforming Beyond-80/20-8B~\citep{wang2025beyond80/20} ($34.6\%$), Alpha-RL-8B~\citep{cai2025alpha-rl} ($28.3\%$), and RL-ZVP-8B~\citep{le2025rl-zvp} ($24.6\%$). 
Notably, \method{} uses only MATH12K and \texttt{Qwen3-4B-Base}, whereas these baselines train on DAPO-MATH-17K and \texttt{Qwen3-8B-Base}. 
Additional details are provided in~\cref{subapp:baseline}.
Even in this unfavorable setting, \method{} achieves stronger performance, underscoring the importance of fully leveraging existing training prompts via composition.

\subsection{Potential For General Domains}\label{subsec:finding3}

Previously, \method{} considered only math problems,%
In this section, we explore whether it can compose problems across domains. 
Specifically, we sample $q_1$ from the physics subset and $q_2$ from MATH12K, yielding a cross-domain compositional dataset, \datasetphysicsmath{}. 
\revise{We compare against the \textit{Mix Training} and \textit{Math-then-Physics} baselines described in~\cref{subsec:exp_setup}, with details in~\cref{subapp:baseline}.}
As shown in~\cref{tab:cross_topic}, we make the following observations:

\ding{182} \textbf{Adding physics prompts for RL training improves multi-task reasoning performance.}
Both the \textit{Mix Training} and \textit{Math-then-Physics} baselines improve GPQA and MMLU-Pro performance relative to training on MATH12K alone. 
On average, \textit{Mix Training} increases the multi-task average by $0.8\%$, and \textit{Math-then-Physics} yields a larger gain of $2.1\%$. 
Moreover, \textit{Math-then-Physics} can further increase performance on math reasoning tasks, whereas \textit{Mix Training} slightly degrades the math reasoning ability. 
These results suggest that, while incorporating physics data benefits multi-task performance, sequential training (math followed by physics) is more effective than mixed training across topics.
As shown in~\cref{fig:composition}, adding physics prompts (via \textit{Mix Training} or \textit{Math-then-Physics}) consistently improves generalization to \textit{law}, \textit{engineering}, and \textit{chemistry} compared to training solely on math (\textit{Math-Only}). 
Interestingly, training on \dataset{} (\textit{Math-Composition}) also yields generalization beyond the math domain.

\ding{183} \textbf{Composing physics and math problems is more effective than naively combining physics and math prompts.}
\revise{RL training on \datasetphysicsmath{} outperforms all baselines by a large margin. Specifically, our method achieves a $+1.3\%$ gain over \textit{Math-then-Physics} and a $+4.1\%$ gain over training solely on MATH12K on MMLU-Pro. On AIME24, it improves by $+7.1\%$ over \textit{Math-then-Physics} and by $+9.1\%$ over training solely on MATH12K. As shown in~\cref{fig:composition}, \textit{Physics-Math-Composition} consistently delivers the best results on both in-domain subjects (math and physics) and OOD subjects (law, engineering, and chemistry).}
\revise{These results show that, in the evaluated math--physics setting, compositional training can improve both in-domain and OOD performance and motivate broader evaluation across domain pairs.}

\ding{184} \revise{\textbf{Multidisciplinary prompts.}
We additionally construct 10K training prompts from SCP-116K~\citep{lu2025scp116k} and compare composition with mixed RL.
As shown in~\cref{tab:cross_topic}, {\method} improves over Mix RL on all six benchmarks and raises overall accuracy from $22.98\%$ to $25.67\%$, providing evidence for broader domains of scientific reasoning.}

\section{Analysis}\label{sec:analysis}

\subsection{Ablation Study}\label{subsec:ablation}

\textbf{About Candidate Sets $\mathcal{D}_k$.}
For $K=2$, our default \method{} constructs each composed prompt by sampling $q_1$ from a small 20-prompt subset $\mathcal{D}_1$ that is resampled for each $q_2$, while drawing $q_2$ from the full prompt pool $\mathcal{D}_2=\mathcal{D}$; we compare this with two alternatives: \textit{A}) both $\mathcal{D}_1$ and $\mathcal{D}_2$ are small random subsets with $|\mathcal{D}_1|=|\mathcal{D}_2|=500$, and \textit{B}) $\mathcal{D}_1=\mathcal{D}$ while $\mathcal{D}_2$ is a small 20-prompt subset resampled for each $q_1$, keeping the total amount of compositional data approximately matched. 
As shown in~\cref{tab:ablation}, our default setting achieves the best overall performance, outperforming \textit{variant~A} by $+3.4\%$ and \textit{variant~B} by $+2.2\%$, while \textit{variant~A} remains close to the baseline despite using only 1K seed prompts, suggesting the potential of \method{} in limited-data regimes. 
The advantage over \textit{variant~B} suggests that diversity in $\mathcal{D}_2$ is particularly important, likely because the composed prompt $q_{1:2}$ shares the final ground-truth answer with $q_2$ and RLVR verifies only the final answer, so using the full $\mathcal{D}_2=\mathcal{D}$ provides broader answer-level training signals.

\textbf{The Necessity of Composition}.
\revise{To validate the necessity of conditioning $q_2$ on $gt_1$, we conduct an ablation study in which $q_1$ and $q_2$ are directly concatenated using the same question pairs as {\method}.
As reported in~\cref{tab:ablation}, this direct-concatenation variant performs even worse than RL training on the original dataset.
Moreover, {\method} outperforms the direct-concatenation variant by $5.5\%$ in overall accuracy.
We believe this degradation stems from a distribution shift, as direct concatenation essentially combines two problems into a single prompt.
In contrast, our {\datamethod} guarantees that the compositional prompt functions as ``one prompt''.}

\revise{\textbf{Intermediate-value control}.
The Feeding-$v_1$ control obtains $28.0\%$ overall, below the $30.3\%$ of {\method} under the same 12K setting (\cref{tab:ablation}).
Thus, the gain cannot be explained solely by exposing the model to the intermediate value; requiring the model to recover and use $v_1$ is important.}

\textbf{Effectiveness on Fully Solvable Prompts}.
\revise{%
We start from an intermediate checkpoint of the 4B model and retain only the solve-all prompts from MATH12K, resulting in 7.2K prompts.
We then compare {\method} on this fully solvable subset against a DAPO-style baseline, in which additional trajectories are adaptively sampled until an incorrect answer is obtained, with a maximum of 32 rollouts.
As shown in~\cref{tab:ablation}, this DAPO-style baseline improves the initial model by only a small margin ($+0.4\%$ overall).
In comparison, {\method} yields a $+2.7\%$ improvement even on this fully solvable subset, even with 8 rollouts per prompt.
These results show that {\method} remains effective when applied from an intermediate checkpoint, even when the retained prompts are relatively simple.}

\subsection{Why \method{} Works}\label{subsec:reasons}

\begin{figure*}[t]
    \centering
    \includegraphics[width=0.45\linewidth]{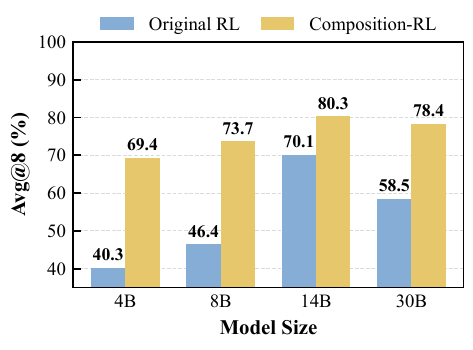}
    \includegraphics[width=0.45\linewidth]{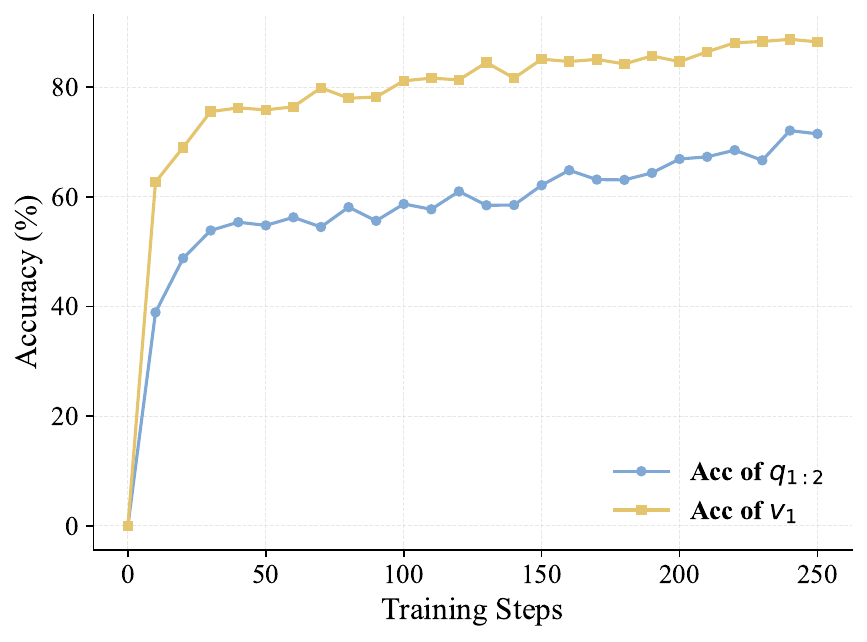}
    \vspace{-0.5em}
    \caption{\revise{\textbf{Left}: \texttt{avg@8} accuracy on Depth-3 compositional test prompts across different model sizes.
    The bars compare Original RL with Composition-RL, i.e., whether Depth-2 compositional prompts are used during RL training.}
\textbf{Right}: The fraction of prompts for which $q_{1:2}$ is solved correctly, and the accuracy of recovering $v_1$ at each training step.}
    \label{fig:analysis}
    \vspace{-0.5em}
\end{figure*}

In this section, we further investigate why \method{} works.
We analyze it from 2 perspectives:

\textbf{Compositional Generalization.}
\textbf{Compositional data may incentivize the acquisition of new skills.}
\citep{yuan2025rl-composition} show that, in a controlled synthetic setting, training on compositional data can elicit new reasoning skills. 
Analogously, if we view an original problem as requiring a stack of skills, composing prompts can create training instances that demand skill recombination. 
\revise{As shown in~\cref{fig:analysis} (left), {\method} substantially improves performance on Depth-3 compositional test prompts: for the 4B, 8B, 14B, and 30B models, \texttt{avg@8} increases from $40.27/46.42/70.09/58.49\%$ without composition to $69.39/73.73/80.35/78.38\%$ with composition.}
\revise{This result is consistent with cross-depth compositional generalization; together with the standard-test gains in~\cref{tab:main_result}, it suggests that the acquired skills transfer beyond the training compositions.}

\textbf{Implicit Process Supervision.}
\textbf{The final-outcome reward for composed prompts also provides implicit signals for the solution process.}
As shown in~\cref{fig:composition}, to solve the composed prompt $q_{1:2}$, LLMs must first obtain $v_1$ and then use $v_1$ to solve $\bar{q}_2$. 
We posit that this structured dependency nudges the model toward a correct intermediate step, at least ``halfway'' through the reasoning. 
As illustrated in~\cref{fig:analysis} (right), the steady improvement in recovering $v_1$ provides evidence that composed prompts can serve as implicit process supervision, even when training relies only on the verification of the final answers.
\revise{Across the analyzed rollouts, only $0.43\%$ reach the correct final answer while recovering an incorrect $v_1$, indicating that successful outcomes rarely bypass the intended intermediate dependency.}
\revise{For more analysis, please refer to~\cref{app:more_analyis}.}

\section{Conclusion \& Discussion}\label{sec:conclusion}

In this paper, we study how to maximize the utility of existing prompts for RL training. 
Comprehensive experiments across various model sizes show that \method{} consistently outperforms RL on the original prompts. 
We also demonstrate the potential of composing prompts from different topics.
Our analysis suggests that compositional prompts can provide implicit process supervision by encouraging correct intermediate steps. 
\revise{Our code, compositional datasets, and trained models are publicly available to support future RL research.}
Promising future directions include:
\ding{182} Extending beyond MATH12K by composing more challenging math training set like Polaris-53K.
\ding{183} Expanding composition to cover more domains.
\ding{184} Adapting \method{} to on-policy distillation~\citep{lu2025onpolicydistillation}.

\section*{\revise{Acknowledgments}}
\revise{This work was partially supported by a grant from the Research Grants Council of the Hong Kong Special Administrative Region, China (Project Reference Number: AoE/E-601/24-N). We would like to thank all reviewers for their helpful suggestions in improving this paper.}
\bibliographystyle{unsrt}
\bibliography{example_paper}

\newpage
\appendix

\section*{Limitation}

\revise{Most experiments use Depth~2, and evidence at higher depths is currently limited to Depth~3. The datasets are constructed offline from a fixed prompt pool; a simple adaptive variant performs on par with the standard construction, so stronger adaptive composition remains future work. Our construction and LLM-based self-verification follow NCSP~\citep{xiao2025ncsp} and therefore inherit its reliance on automated judgments; without a systematic human audit, some invalid prompts may remain after filtering. The method requires $q_1$ to yield an extractable numeric answer and $q_2$ to contain a replaceable numeric condition, although neither source prompt must itself be mathematical; extension to purely textual reasoning and unverifiable domains remains open. Finally, released data or models could be misused; we document dataset construction and intended use to mitigate this risk.}

\section*{Impact Statement}

This paper presents \method{}, which aims to advance research on RLVR. 
\revise{Our two compositional datasets, \dataset{} and \datasetphysicsmath{}, together with our code and trained models, are publicly available resources for future work on RL for LLMs.}
There are many potential societal consequences of our work, none of which we feel must be specifically highlighted here.

\section*{Declaration of LLM Usage}
In this paper, we use \texttt{Qwen2.5-32B-Instruct}~\citep{qwen2.5} to generate our training data for {\method} and \texttt{Qwen3-Base}~\citep{yang2025qwen3} models as the initial models for RL training. 
We also use \texttt{GPT-5.4} to polish the wordings and check for syntax.

\section{Details of Meta-Experiments}\label{app:meta-exp}

\subsection{The \solveall{} Ratio}\label{subapp:meta-exp1}

The training setup in \cref{fig:meta-exp} (Left) follows the main experimental protocol (see \cref{subapp:training}). 
We define the \solveall{} ratio as the fraction of \solveall{} prompts among all over-sampled prompts collected during dynamic-sampling rollouts at a given training step.
For compositional data construction, please refer to~\cref{app:data_composition}.
For RL training details with compositional data, please refer to~\cref{subapp:training} and~\cref{subapp:baseline}.

\subsection{Details of Initial Evaluation For {\datamethod{}}}\label{subapp:meta-exp2}

For the evaluation in \cref{fig:meta-exp} (Right), we randomly sample 200 questions from MATH500 as seed prompts and use \datamethod{} to construct level-2 compositional prompts. 
Specifically, we form pairs $(q_1, q_2)$ by sampling 5 seed questions as candidates for $q_1$ and pairing each with the 200 seed questions as $q_2$. 
After filtering, this procedure yields approximately 400 compositional test prompts. 
We use the same decoding settings as in \cref{subapp:evaluation}.

\section{Experimental Details}\label{app:exp_detail}

\subsection{Training Details}\label{subapp:training}

In \cref{subsec:exp_setup}, we briefly describe the training setup. This appendix provides additional details on our RL training configuration. Two settings are particularly important: we enable dynamic sampling~\citep{yu2025dapo} to filter uninformative prompts, and we use rollout correction to mitigate training--inference mismatch~\citep{yao2025tis}.

\paragraph{Hyperparameters and Rollout Settings.}
Unless otherwise specified, we use a unified set of hyperparameters: batch size $256$ (\revise{rollout batch size is identical to the mini-batch size}), learning rate $1\times 10^{-6}$, and no warm-up. We also adopt a unified rollout configuration: temperature $1$, \texttt{top\_p} $1$, \texttt{top\_k} $-1$, 8 rollouts per problem, and a maximum output length of 16K tokens.

\paragraph{Datasets and Verifiers.}
For the main experiments in \cref{subsec:finding1} and \cref{subsec:finding2}, we train on the MATH training set~\citep{hendrycks2021math}. 
Following the standard protocol, we exclude the MATH500 test set, leaving roughly 12K training prompts spanning five difficulty levels. 
For the cross-topic experiments in \cref{subsec:finding3}, we utilize the physics subset of MegaScience~\citep{fan2025megascience}, which comprises approximately 23K prompts.

For training efficiency, we use Math-Verify as the verifier. 
Considering that rule-based verifiers do not reliably evaluate model outputs on physics problems~\citep{xu2025ugphysics}, we filter the MegaScience physics subset by removing examples for which all eight responses from \texttt{Qwen3-4B-Thinking-2507} are judged incorrect by Math-Verify. 
This yields approximately 8.2K prompts on which rule-based verification is reliable.

\subsection{Baselines}\label{subapp:baseline}

In this appendix, we provide additional baseline details for the experiments in~\cref{subsec:finding1,subsec:finding2,subsec:finding3}.

For the experiments in~\cref{subsec:finding1}, the baseline corresponds to RL training on the original MATH12K training set, which contains 12K training prompts. 
Our \method{} trains instead on compositional prompts constructed from MATH12K. 
\revise{As described in~\cref{subsec:method_rl_composition}, for each $q_2$ drawn from the full dataset, we independently resample a uniform 20-prompt subset from which $q_1$ is drawn, yielding $20\times 12\text{K}=240\text{K}$ compositional prompts in principle.}
As discussed in~\cref{subapp:reliability}, we apply a verification-and-filtering procedure to improve data quality. 
After verification of Step 1, approximately 231K prompts remain; after verifying Step 2, this is reduced to roughly 200K; and after the final check, we obtain about 199K compositional prompts.
We refer to this composition set as \dataset{}.

For the experiments in~\cref{subsec:finding2}, we include Beyond-80/20~\citep{wang2025beyond80/20}, AlphaRL~\citep{cai2025alpha-rl}, and RL-ZVP~\citep{le2025rl-zvp} as additional reference baselines. 
We use the models that are initialized from \texttt{Qwen3-8B-Base} and trained on DAPO-MATH-17K prompts from these works. 
We report their results as RL-zero baselines to our curriculum-based \method{} trained from \texttt{Qwen3-4B-Base}. 
This comparison is unfavorable to \method{} due to differences in both model scale and training data.
For curriculum \method{}, we first train \texttt{Qwen3-4B-Base} on the original MATH12K set (Depth~1). 
After performance saturates, we switch to the Depth~2 training set (\dataset{}), and then to the Depth~3 training set once Depth~2 saturates. 
The construction of the Depth~3 compositional set follows the procedure used for Depth~2.
Since Beyond-80/20~\citep{wang2025beyond80/20}, AlphaRL~\citep{cai2025alpha-rl}, and RL-ZVP~\citep{le2025rl-zvp} have not released their models at the time we were writing our paper, we report their results as quoted directly from the corresponding papers.

For the experiments in~\cref{subsec:finding3}, we consider two natural RL baselines. 
The first is RL training on a mixture of MATH12K and the MegaScience Physics subset, which we denote as \textit{Mix Training}. 
The second baseline continues RL training on Physics data, starting from a checkpoint trained on MATH12K, which we denote as \textit{Math-then-Physics}. 
For \textit{Math-then-Physics}, we train until performance saturates. 
For a fair comparison, we train \textit{Mix Training} for approximately the same number of total gradient updates as the combined MATH12K stage plus the physics training stage.
\revise{For \method{}, we sample $q_1$ from Physics and $q_2$ from MATH12K.}
\revise{After filtering, the resulting compositional dataset contains approximately 141K prompts, which we denote as \datasetphysicsmath{}.}

\subsection{Evaluation Details}\label{subapp:evaluation}

To comprehensively evaluate model capabilities, we use a diverse suite of benchmarks spanning mathematical reasoning and multi-task reasoning:

\begin{enumerate}
    \item \textbf{Mathematical reasoning:} We evaluate on AIME24, AIME25, BeyondAIME~\citep{bytedance_seed_2025_beyondaime}, and IMO-Bench~\citep{luong2025imobench}. 
    Since AIME24 and AIME25 each contain 30 problems, we report \texttt{pass@1} using 32 samples per problem (\texttt{avg@32}). 
    BeyondAIME contains 100 problems; we report \texttt{avg@8}. 
    For IMO-Bench, we use the AnswerBench subset to enable rule-based verification; it contains 400 problems, and we report \texttt{avg@4}.

    \item \textbf{Multi-task reasoning:} We evaluate on GPQA-Diamond~\citep{rein2024gpqa} (approximately 200 problems) and report \texttt{pass@1} using 8 samples per problem. 
    We also evaluate on MMLU-Pro~\citep{wang2024mmlupro}; since it contains over 5K problems, we report results from a single run.
\end{enumerate}

All evaluation codes are adapted from the DeepscaleR~\citep{deepscaler2025} codebase, and we use vLLM~\citep{kwon2023vllm} to accelerate inference and Math-Verify to evaluate the LLMs' answers. For decoding, we \revise{follow~\citep{xu2025thinking}} and set the temperature to 0.6, \texttt{top\_p} to 0.95, \texttt{top\_k} to 20, and the maximum output length to 32K tokens.

\subsection{Details of Ablation for Candidate Sets $\mathcal{D}_k$}\label{subapp:ablation}

As noted in~\cref{subsec:ablation}, the default configuration of \method{} sets $\mathcal{D}_2=\mathcal{D}$ (the full prompt pool) and samples $\mathcal{D}_1$ as a small random subset ( $|\mathcal{D}_2|=12{,}000$, $|\mathcal{D}_1|=20$ ).  
We also consider two variants:  
A) Both $\mathcal{D}_1$ and $\mathcal{D}_2$ are small randomly sampled subsets ( $|\mathcal{D}_1|=|\mathcal{D}_2|=500$ ).  
B) $\mathcal{D}_1$ is the full set $\mathcal{D}$, while $\mathcal{D}_2$ is a small randomly sampled subset ( $|\mathcal{D}_1|=12{,}000$, $|\mathcal{D}_2|=20$ ).  
These settings are designed to yield roughly the same theoretical compositional dataset size. 

After applying the filtering procedure in~\cref{subapp:reliability}, the resulting actual dataset sizes are:  
\begin{itemize}
    \item \method{}: 199K (see~\cref{subapp:baseline}).  
    \item \textit{Variant~A}: 240K after step 1, 202K after step 2, and 200K after step 3.
    \item \textit{Variant~B}: 231K after step 1, 201K after step 2, and 200K after step 3.
\end{itemize}
Thus, the final dataset sizes are approximately matched across configurations.

\section{Analysis Details}\label{app:analyis}

In this appendix, we provide additional details for~\cref{subsec:reasons}.

To evaluate compositional generalization, we use the same setting as in~\cref{subapp:meta-exp1}. 
To determine whether the first variable $v_1$ is solved correctly, we prompt \texttt{Qwen2.5-32B-Instruct} using the default generation configuration and the prompt shown in~\cref{appfig:v1_verify_prompt}.

\begin{figure}[htbp!]
\centering

\begin{tcolorbox}[
  title=\textbf{Prompt for Verifying the Correctness of Finding $v_1$ in LLMs' Response},
  coltitle=white,
  width=\linewidth,
  halign title=center,
  left=3pt,right=3pt,top=2pt,bottom=2pt
]
\ttfamily\small
You are a math solution verifier. Your task is to check if a given response correctly solves for a specific intermediate variable in a composite math problem.

\medskip
**Problem:**
\{problem\}

\medskip
**Target Variable:** \$\{symbol\}\$\\
**Variable Definition:** \{definition\}\\
**Correct Answer for \$\{symbol\}\$:** \{correct\_answer\}

\medskip
**Model's Response:**
\{response\}

\medskip
---

\medskip
**Your Task:**
1. Identify what value the model computed for \$\{symbol\}\$ in its response.\\
2. Compare it with the correct answer: \{correct\_answer\}.\\
3. Determine if the model correctly solved for \$\{symbol\}\$.

\medskip
**Important Notes:**
- Focus ONLY on whether \$\{symbol\}\$ was correctly computed; ignore the final answer of the composite problem.\\
- The value might be stated explicitly (e.g., ``\$\{symbol\}\$ = 7'') or implicitly derived.\\
- Accept equivalent forms (e.g., ``7'', ``7.0'', ``seven'' are all correct if the answer is 7).

\medskip
**Output in JSON format:**
\{\\
\ \ "extracted\_value": "<the value the model gave for \{symbol\}, or 'NOT\_FOUND' if not mentioned>",\\
\ \ "is\_equivalent": <true if extracted\_value equals \{correct\_answer\}, false otherwise>,\\
\ \ "reasoning": "<brief explanation>",\\
\ \ "verdict": "<CORRECT or INCORRECT>"\\
\}
\normalfont
\end{tcolorbox}

\caption{The Prompt for Verifying the Correctness of Finding $v_1$ in LLMs' Response.}
\label{appfig:v1_verify_prompt}
\end{figure}

\section{More Analysis}\label{app:more_analyis}

\subsection{\revise{Difficulty-Stratified Compositional Evaluation}}

\revise{We stratify the evaluation pairs by the difficulty of their first and second source problems.  In~\cref{tabapp:difficulty_composition}, ``N'' and ``Y'' denote evaluation without and with composition, respectively.  Composing the sources consistently reduces \solveall{} cases and increases informative cases, including when both sources are individually easy.}

\begin{table*}[t]
\centering
\small
\caption{\revise{Difficulty-stratified evaluation (\%).  ``Informative'' denotes prompts that are neither \solveall{} nor \solvenone{}.}}
\label{tabapp:difficulty_composition}
\begin{tabular}{llccccc}
\toprule
\revise{\textbf{Source difficulty}} & \revise{\textbf{Composition}} & \revise{\textbf{pass@1}} & \revise{\textbf{pass@8}} & \revise{\textbf{\solveall{}}} & \revise{\textbf{\solvenone{}}} & \revise{\textbf{Informative}} \\
\midrule
\revise{Easy$\rightarrow$Easy} & \revise{N} & \revise{99.3} & \revise{100.0} & \revise{94.5} & \revise{0.0}  & \revise{5.5}  \\
\revise{Easy$\rightarrow$Easy} & \revise{Y} & \revise{89.2} & \revise{98.6}  & \revise{69.9} & \revise{1.4}  & \revise{28.8} \\
\midrule
\revise{Easy$\rightarrow$Hard} & \revise{N} & \revise{66.3} & \revise{78.3} & \revise{51.7} & \revise{21.7} & \revise{26.7} \\
\revise{Easy$\rightarrow$Hard} & \revise{Y} & \revise{57.7} & \revise{83.3} & \revise{26.7} & \revise{16.7} & \revise{56.7} \\
\midrule
\revise{Hard$\rightarrow$Easy} & \revise{N} & \revise{99.8} & \revise{100.0} & \revise{98.3} & \revise{0.0} & \revise{1.7}  \\
\revise{Hard$\rightarrow$Easy} & \revise{Y} & \revise{68.5} & \revise{93.1}  & \revise{34.5} & \revise{6.9} & \revise{58.6} \\
\midrule
\revise{Hard$\rightarrow$Hard} & \revise{N} & \revise{81.2} & \revise{94.0} & \revise{64.2} & \revise{6.0}  & \revise{29.9} \\
\revise{Hard$\rightarrow$Hard} & \revise{Y} & \revise{47.2} & \revise{73.1} & \revise{20.9} & \revise{26.9} & \revise{52.2} \\
\bottomrule
\end{tabular}
\end{table*}

\subsection{\revise{Overlap of Fully Solved Source Prompts}}

\revise{To reconcile the increasing fraction of fully solved source prompts with the continued utility of composition, we measure how often the two sources of a composed prompt are already \solveall{}.  As training proceeds, most pairs contain at least one \solveall{} source, and by step 250 both sources are \solveall{} for 69.0\% of pairs.  Composition can therefore create informative training instances even from source prompts that have largely saturated in isolation.}

\begin{table}[t]
\centering
\small
\caption{\revise{Overlap of \solveall{} source prompts during training (\%).}}
\label{tabapp:solveall_overlap}
\begin{tabular}{ccc}
\toprule
\revise{\textbf{Step}} & \revise{\textbf{At least one source}} & \revise{\textbf{Both sources}} \\
 & \revise{\textbf{is \solveall{}}} & \revise{\textbf{are \solveall{}}} \\
\midrule
\revise{50}  & \revise{84.8} & \revise{36.4} \\
\revise{100} & \revise{91.2} & \revise{48.7} \\
\revise{200} & \revise{95.7} & \revise{62.1} \\
\revise{250} & \revise{97.3} & \revise{69.0} \\
\bottomrule
\end{tabular}
\end{table}

\paragraph{Further evidence of implicit process supervision}

We provide more evidence of our ``implicit process supervision'' claim based on AIME24 using the 4B model. We consider three metrics:
(i) \textbf{First-Attempt Correctness}, defined as the proportion of responses that produce the correct answer with a single \verb|\boxed{}|;
(ii) \textbf{Reflection Effectiveness}, defined as the accuracy conditioned on the presence of self-correction keywords; and
(iii) \textbf{Self-Correction Success Rate}, defined as the correctness when the model revises its \verb|\boxed{}| answer.

\revise{As shown in Table~\ref{tab:implicit_process_supervision}, our method consistently outperforms the baseline on all three metrics.
Specifically, first-attempt correctness improves from 42.8 to 51.7, indicating that our method leads to stronger initial reasoning before any explicit revision.
Moreover, the accuracy conditioned on reflection increases from 11.0 to 19.8, suggesting that reflective behavior is more effective under our method (implicit process supervision).
Finally, the self-correction success rate rises substantially from 9.1 to 26.1, showing that when the model revises its answer, the revision is much more likely to be correct.
Taken together, these results are consistent with implicit process supervision, with improvements in final accuracy, intermediate reasoning, and self-correction behavior.}

\begin{table}[t]
\centering
\small
\begin{tabular}{lcc}
\toprule
\textbf{Metric (\%)} & \textbf{Baseline} & \textbf{Ours} \\
\midrule
First-attempt correctness & 42.8 & 51.7 \\
$P(\text{correct} \mid \text{reflection})$ & 11.0 & 19.8 \\
Self-correction success & 9.1 & 26.1 \\
\bottomrule
\end{tabular}
\caption{Additional evidence of implicit process supervision on AIME24 with the 4B model.}
\label{tab:implicit_process_supervision}
\end{table}

\paragraph{Error analysis.}
\revise{
We further conduct an error analysis on cases where the model can solve both constituent questions in isolation but fails on the corresponding compositional question. We manually inspect 40 such cases and categorize the errors into three types: (1) \textbf{Linkage error}, where the model fails at the ``linkage condition'' for any reasons; (2) \textbf{$q_1$ error}, where the model makes an error on $q_1$ when solving it in the compositional setting, despite being able to solve it correctly in isolation; and (3) \textbf{$q_2$ error}, where the model solves $q_1$ and derives the linkage correctly but still fails on $q_2$. As shown in Table~\ref{tab:error_analysis_cases}, linkage error is the dominant failure mode for the original model, accounting for 22 out of 40 cases, compared with 7 $q_1$ errors and 11 $q_2$ errors. After applying {\method}, all three error types are reduced, with the largest reduction observed in linkage errors (from 22 to 7). Qualitatively, we find that linkage errors often arise from incorrect intermediate computation, unproductive self-verification loops, or selecting an incorrect linkage value despite otherwise correct reasoning.
Interestingly, these examples also shed light on common failure modes of current reasoning models: (1) intermediate computation errors during the thinking process, (2) repeated self-verification loops that lead to truncation without progress, and (3) loss of direction mid-reasoning even when correct intermediate results are available. Composition-RL reduces linkage errors across all three failure modes, suggesting an improved ability to maintain correct intermediate reasoning throughout multi-step problems.
}

\begin{table}[t]
\centering
\small
\begin{tabular}{lcc}
\toprule
\textbf{Error Type} & \textbf{Original} & \textbf{Composition-RL} \\
\midrule
Linkage error & 22 & 7 \\
$q_1$ error   & 7  & 1 \\
$q_2$ error   & 11 & 5 \\
\bottomrule
\end{tabular}
\caption{Error analysis on 40 cases where the model solves both constituent questions in isolation but fails on the compositional question.}
\label{tab:error_analysis_cases}
\end{table}

\section{More Results}\label{app:more_results}

\subsection{About Average Response Length}

\revise{
In~\cref{sec:exp}, we primarily discuss results based on \texttt{pass@1}; here, we also report the average response length in~\cref{tabapp:avg_length}.
As shown in~\cref{tabapp:avg_length}, {\method} does not significantly increase response length across all model scales.
}

\begin{table*}[!t]
    \centering
    \caption{Average response length (in K tokens) across different benchmarks. 
    ``ID Avg'' denotes the average response length over in-domain mathematics benchmarks, 
    ``OOD Avg'' denotes the average response length over out-of-domain multi-task benchmarks, 
    and ``Avg'' denotes the overall average.}
    \vspace{-1em}
    \label{tabapp:avg_length}
    \vspace{2mm}
    \resizebox{0.98\textwidth}{!}{
\begin{tabular}{c|ccccc|ccc|c}
\toprule[1.6pt]
    \multirow{2}{*}{\textbf{Composition}} 
    & \multicolumn{5}{c|}{\textbf{Mathematics (In-Domain)}} 
    & \multicolumn{3}{c|}{\textbf{Multi-Task (Out-Of-Domain)}}
    & \textbf{Overall} \\ \cmidrule(lr){2-10}
    ~ & \textbf{AIME 24} 
      & \textbf{AIME 25} 
      & \textbf{Beyond AIME} 
      & \textbf{IMO-Answer}
      & \textbf{ID Avg}
      & \textbf{GPQA}
      & \textbf{MMLU-Pro}
      & \textbf{OOD Avg}
      & \textbf{Avg} \\ 
\midrule
\multicolumn{10}{c}{\texttt{\textbf{Qwen3-4B-Base}}} \\
\midrule
\ding{56} & 16.0 & 13.4 & 13.3 & 12.8 & 13.9 & 4.2 & 2.3 & 3.3 & 10.3 \\
\rowcolor[rgb]{ .867, .922, .969}
\ding{52} & 14.7 & 13.2 & 12.3 & 12.9 & 13.3 & 4.0 & 2.0 & 3.0 & 9.9 \\
\midrule
\multicolumn{10}{c}{\texttt{\textbf{Qwen3-8B-Base}}} \\
\midrule
\ding{56} & 17.5 & 15.7 & 16.9 & 15.8 & 16.5 & 5.1 & 2.5 & 3.8 & 12.3 \\
\rowcolor[rgb]{ .867, .922, .969}
\ding{52} & 14.2 & 14.9 & 15.9 & 14.9 & 15.0 & 5.5 & 2.8 & 4.2 & 11.4 \\
\midrule
\multicolumn{10}{c}{\texttt{\textbf{Qwen3-14B-Base}}} \\
\midrule
\ding{56} & 13.6 & 13.4 & 12.9 & 12.1 & 13.0 & 3.7 & 1.9 & 2.8 & 9.6 \\
\rowcolor[rgb]{ .867, .922, .969}
\ding{52} & 15.3 & 14.3 & 15.9 & 15.9 & 15.4 & 4.9 & 2.2 & 3.6 & 11.4 \\
\midrule
\multicolumn{10}{c}{\texttt{\textbf{Qwen3-30B-A3B-Base}}} \\
\midrule
\ding{56} & 13.0 & 16.1 & 15.3 & 16.2 & 15.2 & 4.4 & 2.3 & 3.4 & 11.2 \\
\rowcolor[rgb]{ .867, .922, .969}
\ding{52} & 13.1 & 15.6 & 14.5 & 15.3 & 14.6 & 4.3 & 2.1 & 3.2 & 10.8 \\
\bottomrule[1.6pt]
\end{tabular}}
\vspace{-0.5em}
\end{table*}

\subsection{About \texttt{pass@k}}

\revise{In~\cref{sec:exp}, we primarily discuss results based on \texttt{pass@1}; here, we additionally report \texttt{pass@k} results in~\cref{tabapp:passk_result}.
As shown in~\cref{tabapp:passk_result}, {\method} consistently improves \texttt{pass@k} across model scales, indicating that it broadens the set of solvable problems.}

\begin{table*}[!t]
    \centering
    \caption{Pass@k (\%) across different benchmarks. 
    The value of $k$ for each benchmark is consistent with that used in the corresponding Avg@k setting in the main results table.
    ``ID Avg'' denotes the average pass@k over in-domain mathematics benchmarks, 
    ``OOD Avg'' denotes the average pass@k over out-of-domain multi-task benchmarks, 
    and ``Avg'' denotes the overall average.}
    \vspace{-1em}
    \label{tabapp:passk_result}
    \vspace{2mm}
    \resizebox{0.98\textwidth}{!}{
\begin{tabular}{c|ccccc|ccc|c}
\toprule[1.6pt]
    \multirow{3}{*}{\textbf{Composition}} 
    & \multicolumn{5}{c|}{\textbf{Mathematics (In-Domain)}} 
    & \multicolumn{3}{c|}{\textbf{Multi-Task (Out-Of-Domain)}}
    & \textbf{Overall} \\ \cmidrule(lr){2-10}
    ~ & \textbf{AIME 24} 
      & \textbf{AIME 25} 
      & \textbf{Beyond AIME} 
      & \textbf{IMO-Answer}
      & \textbf{ID Avg}
      & \textbf{GPQA}
      & \textbf{MMLU-Pro}
      & \textbf{OOD Avg}
      & \textbf{Avg} \\ 
    & \texttt{Pass@32} & \texttt{Pass@32} & \texttt{Pass@8} & \texttt{Pass@4}
      & \texttt{Avg.}
      & \texttt{Pass@8} & \texttt{Pass@1}
      & \texttt{Avg.}
      & \texttt{Avg.} \\ 
\midrule
\multicolumn{10}{c}{\texttt{\textbf{Qwen3-4B-Base}}} \\
\midrule
\ding{56} & 56.7 & 46.7 & 23.0 & 25.5 & 38.0 & 79.3 & 58.6 & 69.0 & 48.3 \\
\rowcolor[rgb]{ .867, .922, .969}
\ding{52} & 66.7 & 56.7 & 31.0 & 26.0 & 45.1 & 77.8 & 61.4 & 69.6 & 53.3 \\
\midrule
\multicolumn{10}{c}{\texttt{\textbf{Qwen3-8B-Base}}} \\
\midrule
\ding{56} & 66.7 & 46.7 & 29.0 & 30.7 & 43.3 & 81.3 & 62.6 & 72.0 & 52.8 \\
\rowcolor[rgb]{ .867, .922, .969}
\ding{52} & 73.3 & 53.3 & 35.0 & 32.0 & 48.4 & 81.3 & 64.5 & 72.9 & 56.6 \\
\midrule
\multicolumn{10}{c}{\texttt{\textbf{Qwen3-14B-Base}}} \\
\midrule
\ding{56} & 70.0 & 70.0 & 39.0 & 34.5 & 53.4 & 81.8 & 67.2 & 74.5 & 60.4 \\
\rowcolor[rgb]{ .867, .922, .969}
\ding{52} & 76.7 & 76.7 & 39.0 & 41.5 & 58.5 & 81.8 & 69.3 & 75.6 & 64.2 \\
\midrule
\multicolumn{10}{c}{\texttt{\textbf{Qwen3-30B-A3B-Base}}} \\
\midrule
\ding{56} & 73.3 & 56.7 & 39.0 & 40.0 & 52.3 & 83.8 & \revise{62.6} & \revise{73.2} & \revise{59.2} \\
\rowcolor[rgb]{ .867, .922, .969}
\ding{52} & 76.7 & 63.3 & 40.0 & 39.0 & 54.8 & 84.8 & \revise{64.6} & \revise{74.7} & \revise{61.4} \\
\bottomrule[1.6pt]
\end{tabular}}
\vspace{-0.5em}
\end{table*}

\subsection{\revise{Variation Across Training Seeds}}

\revise{We repeat the 30B-A3B baseline and \method{} training runs with three random seeds.  Table~\ref{tabapp:seed_std} reports the standard deviation across seeds; the corresponding mean scores are reported in~\cref{tab:main_result}.  The small and comparable deviations indicate that the gains are stable across the evaluated seeds.}

\begin{table}[t]
    \centering
    \small
    \caption{\revise{Standard deviation of \texttt{pass@1} (percentage points) over three training seeds for \texttt{Qwen3-30B-A3B-Base}.}}
    \label{tabapp:seed_std}
    \begin{tabular}{lcc}
    \toprule
    \revise{\textbf{Benchmark}} & \revise{\textbf{Baseline}} & \revise{\textbf{\method{}}} \\
    \midrule
    \revise{AIME24}     & \revise{1.58} & \revise{1.65} \\
    \revise{AIME25}     & \revise{1.52} & \revise{1.60} \\
    \revise{BeyondAIME} & \revise{1.23} & \revise{1.35} \\
    \revise{GPQA}       & \revise{1.08} & \revise{1.14} \\
    \revise{IMOBench}   & \revise{1.27} & \revise{1.33} \\
    \revise{MMLU-Pro}   & \revise{0.80} & \revise{0.92} \\
    \bottomrule
    \end{tabular}
\end{table}

\subsection{\revise{Generalization to a Non-Qwen Backbone}}

\revise{We additionally evaluate the baseline and \method{} with the non-Qwen \texttt{OctoThinker-3B} backbone~\citep{wang2025octothinker}. As shown in~\cref{tabapp:octothinker}, \method{} improves all nine benchmarks, indicating that its benefits are not specific to the Qwen model family.}

\begin{table}[t]
    \centering
    \small
    \caption{\revise{\texttt{pass@1} (\%) with \texttt{OctoThinker-3B}.}}
    \label{tabapp:octothinker}
    \begin{tabular}{lcc}
    \toprule
    \revise{\textbf{Benchmark}} & \revise{\textbf{Baseline}} & \revise{\textbf{\method{}}} \\
    \midrule
    \revise{AIME24}         & \revise{3.44}  & \revise{5.94}  \\
    \revise{AIME25}         & \revise{1.88}  & \revise{2.81}  \\
    \revise{AMC}            & \revise{24.85} & \revise{28.12} \\
    \revise{BeyondAIME}     & \revise{1.25}  & \revise{1.62}  \\
    \revise{GPQA}           & \revise{25.88} & \revise{29.23} \\
    \revise{IMO-Answer}     & \revise{5.31}  & \revise{5.44}  \\
    \revise{MATH}           & \revise{57.60} & \revise{63.15} \\
    \revise{MMLU-Pro}       & \revise{34.55} & \revise{36.66} \\
    \revise{OlympiadBench}  & \revise{21.33} & \revise{22.81} \\
    \bottomrule
    \end{tabular}
\end{table}

\subsection{Ratio of \solveall{} and \solvenone{}}

\begin{figure}
    \centering
    \noindent\raisebox{0.14\linewidth}{\makebox[0.08\linewidth][r]{\revise{\textbf{4B}}\hspace{0.5em}}}%
    \includegraphics[width=0.4\linewidth]{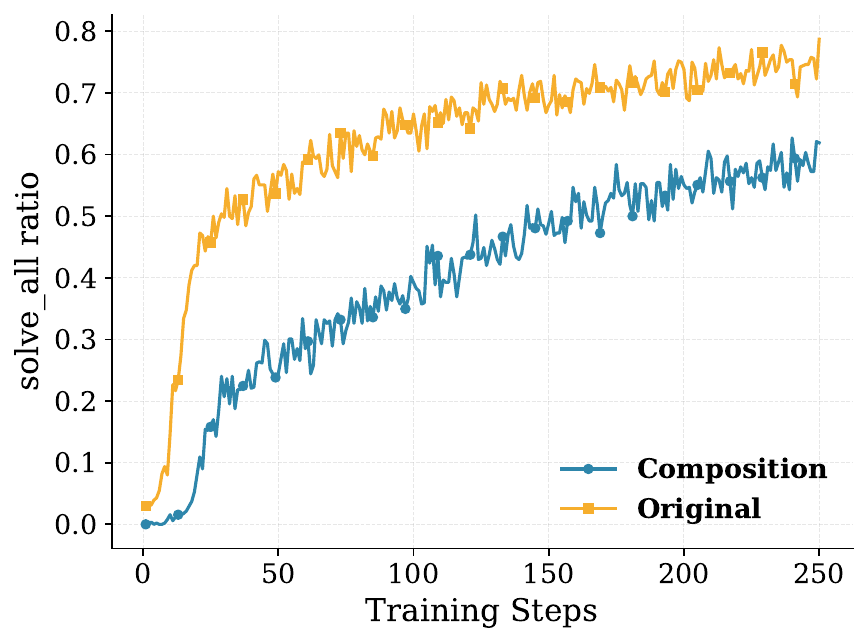}%
    \includegraphics[width=0.4\linewidth]{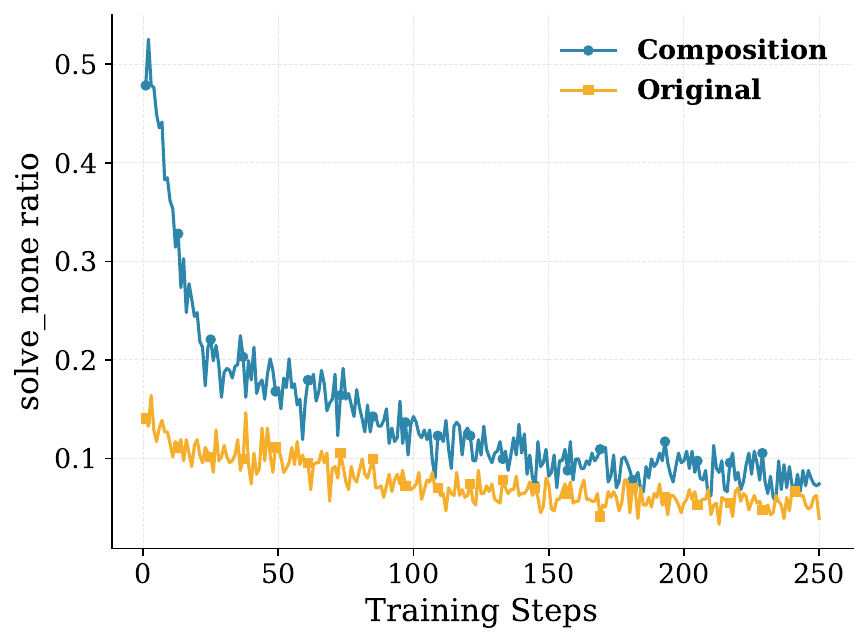}\par
    \noindent\raisebox{0.14\linewidth}{\makebox[0.08\linewidth][r]{\revise{\textbf{8B}}\hspace{0.5em}}}%
    \includegraphics[width=0.4\linewidth]{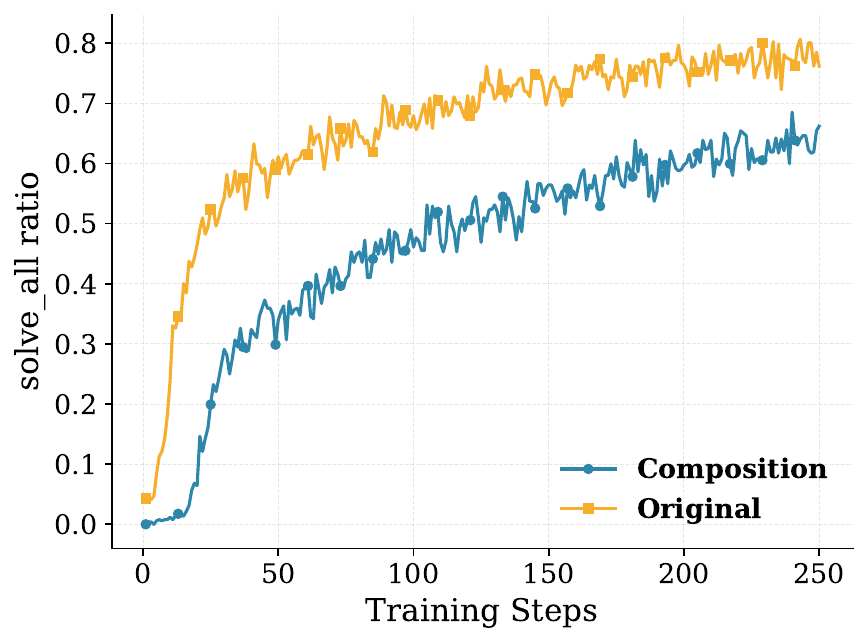}%
    \includegraphics[width=0.4\linewidth]{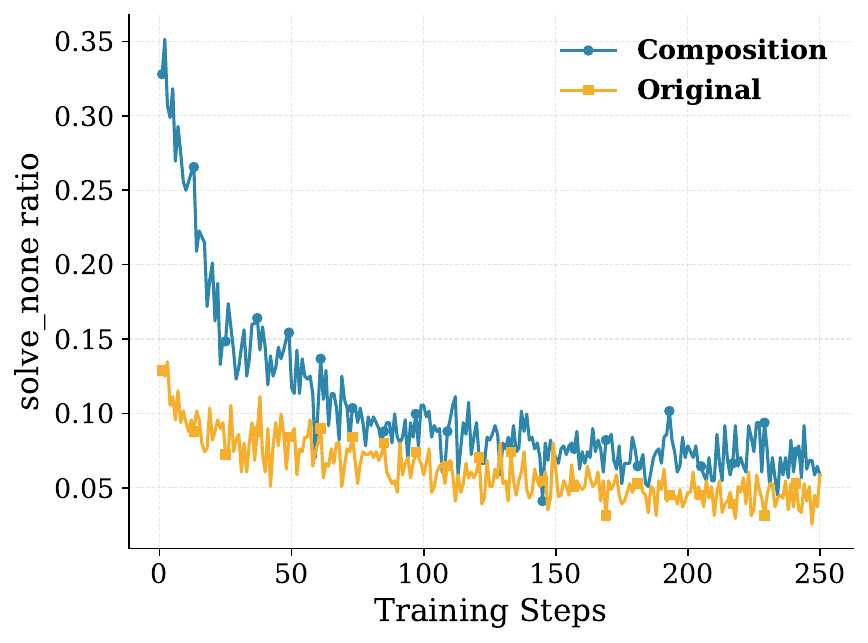}\par
    \noindent\raisebox{0.14\linewidth}{\makebox[0.08\linewidth][r]{\revise{\textbf{14B}}\hspace{0.5em}}}%
    \includegraphics[width=0.4\linewidth]{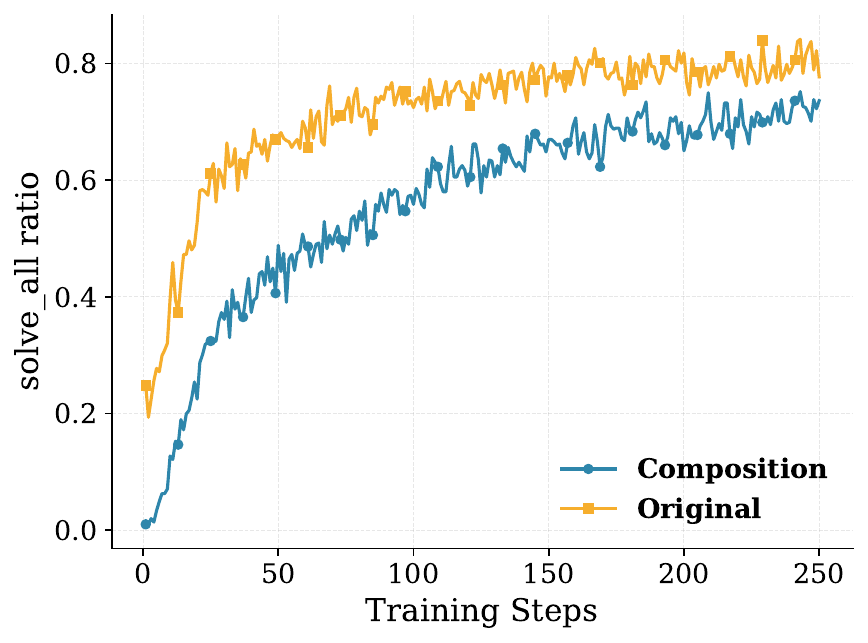}%
    \includegraphics[width=0.4\linewidth]{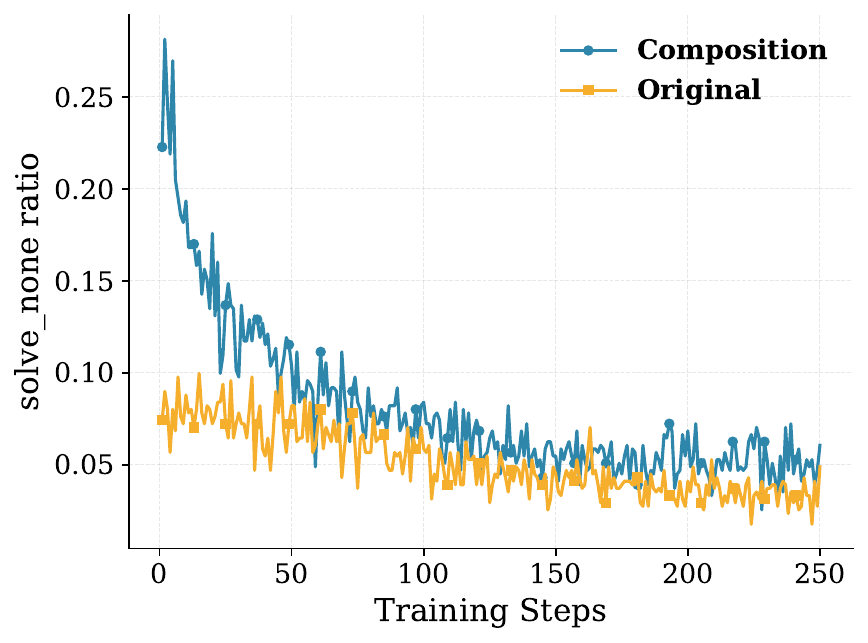}\par
    \noindent\raisebox{0.14\linewidth}{\makebox[0.08\linewidth][r]{\revise{\textbf{30B}}\hspace{0.5em}}}%
    \includegraphics[width=0.4\linewidth]{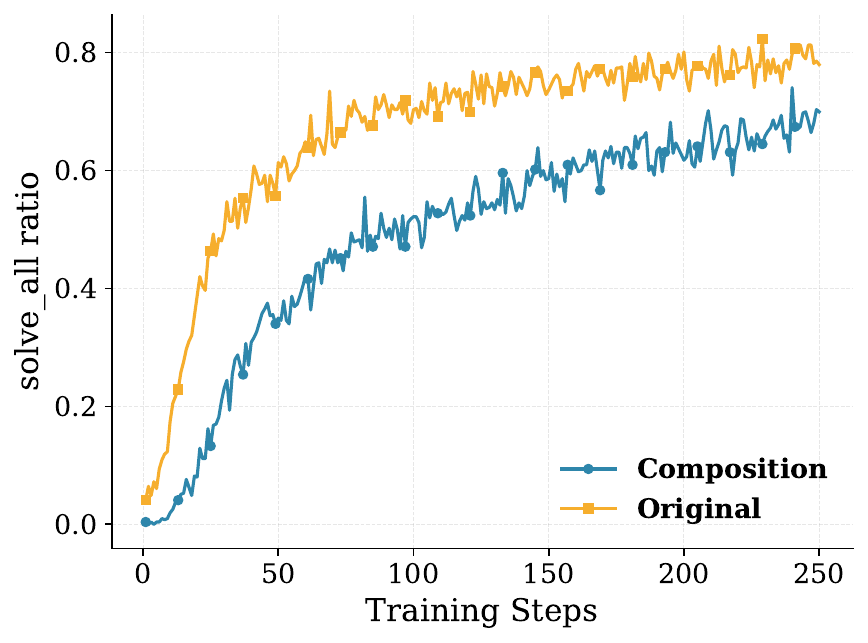}%
    \includegraphics[width=0.4\linewidth]{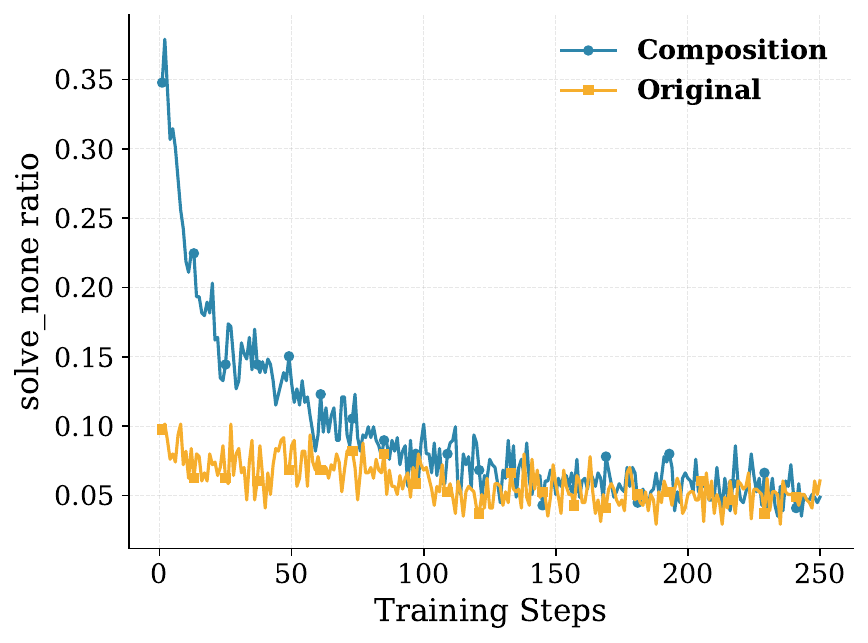}
    \caption{\solveall{} and \solvenone{} ratio across model sizes.}
    \label{fig:solve_all_solve_none}
\end{figure}

\revise{We also report the proportions of \solveall{} and \solvenone{} examples across training steps in~\cref{fig:solve_all_solve_none}.
As shown in~\cref{fig:solve_all_solve_none}, at later stages of training, {\method} exhibits a lower \solveall{} ratio and a comparable \solvenone{} ratio.
Therefore, {\method} yields a larger amount of effective training data overall.}

\section{Details of {\datamethod}}\label{app:data_composition}

\subsection{Reliability of {\datamethod}}\label{subapp:reliability}

The full \datamethod{} pipeline can be automated with an LLM assistant; implementation details and the corresponding prompts are provided in \cref{subapp:compose_detail} and \cref{subapp:compose_prompt}, respectively. 
To make this automated process more reliable, we have some additional verification steps to filter potential mistakes during composing.
Following \cite{xiao2025ncsp}, we use LLM-based self-verification at each composition step. 
Concretely, we prompt the same LLM to perform the following checks:

\begin{itemize}
    \item \ding{182} \textbf{Verification of ``Modify $q_1$ with $gt_1$.''} In this step, the LLM extracts a variable $v_1$ from $gt_1$ and its definition $d_1$. We then ask the LLM to compute the value of $v_1$ given $q_1$ and $d_1$, and compare the computed value against the extracted $v_1$. If they do not match, we discard the prompt. This verification improves the reliability of the modification of $q_1$.
    \item \ding{183} \textbf{Verification of ``Modify $q_2$.''} Analogously, we prompt the LLM to verify whether the extracted variable $v_2$ (and its definition) is consistent with $q_2$. Prompts that fail this check are filtered out.
    \item \ding{184} \textbf{Verification of ``Connect $q_1$ and $q_2$.''} This step primarily involves concatenation. To ensure quality, we prompt the LLM to check for inconsistencies (e.g., conflicting variable names) and filter out any inconsistent prompts.
\end{itemize}

This verification procedure removes many low-quality compositions, leaving a substantially more reliable set of composed prompts. 
As reported in \cite{xiao2025ncsp}, the rate of erroneous prompts after filtering is below $2\%$. 
We believe this error rate is acceptable for training.

\subsection{Implementation Details}\label{subapp:compose_detail}

We use \texttt{Qwen2.5-32B-Instruct}~\citep{qwen2.5} with step-specific prompts to implement each stage of \cref{subsec:spc} as well as the verification procedure in \cref{subapp:reliability}. 
Unless otherwise specified, we set the temperature to $0.1$, $\texttt{top\_p}$ to $0.7$, and the maximum output length to $4096$ tokens. 
The prompts are provided in \cref{subapp:compose_prompt}.

\subsection{\revise{Physics-to-Mathematics Composition Examples}}

\revise{The following examples give both source questions and the complete composed prompt. The answer-derived value from the physics question $q_1$ determines a replacement in the mathematics question $q_2$.}

\begin{tcolorbox}[
  title=\textbf{\revise{\textcolor{white}{Example 1: Mechanics to Complex Numbers}}},
  coltitle=white,
  width=\linewidth,
  halign title=center,
  left=3pt,right=3pt,top=2pt,bottom=2pt
]
\revise{\textbf{First question $q_1$:} If a force of $50\,\mathrm{N}$ is applied to an object with an acceleration of $2\,\mathrm{m\,s^{-2}}$, what is the mass of the object? Its ground-truth answer is $25\,\mathrm{kg}$.}

\medskip
\revise{\textbf{Second question $q_2$:} Let $z_1,z_2,z_3$ be complex numbers such that $|z_1|=2$, $|z_2|=3$, and $|z_3|=4$. Find the largest possible value of $|z_1-z_2|^2+|z_1-z_3|^2+|z_2-z_3|^2$. Its ground-truth answer is $87$.}

\medskip
\revise{\textbf{Composed question $q_{1:2}$:} If a force of $50\,\mathrm{N}$ is applied to an object with an acceleration of $2\,\mathrm{m\,s^{-2}}$, what is the mass of the object? Let $X$ denote the numerical value of the mass in kilograms, and let $n$ be $23$ less than $X$. Let $z_1,z_2,z_3$ be complex numbers such that $|z_1|=n$, $|z_2|=3$, and $|z_3|=4$. Find the largest possible value of $|z_1-z_2|^2+|z_1-z_3|^2+|z_2-z_3|^2$. Here $X=25$ and $n=2$, so the composed prompt preserves the ground-truth answer $87$.}
\end{tcolorbox}

\begin{tcolorbox}[
  title=\textbf{\revise{\textcolor{white}{Example 2: Magnetism to Complex Roots}}},
  coltitle=white,
  width=\linewidth,
  halign title=center,
  left=3pt,right=3pt,top=2pt,bottom=2pt
]
\revise{\textbf{First question $q_1$:} An air-cored coil has a flux density of $25\,\mathrm{mT}$. When an iron core is inserted, the flux density increases to $1.6\,\mathrm{T}$. Calculate the relative permeability of the iron. Its ground-truth answer is $\mu_r=64$.}

\medskip
\revise{\textbf{Second question $q_2$:} Let $x$ be a complex number such that $x^{2011}=1$ and $x\neq1$. Compute
$\frac{x^2}{x-1}+\frac{x^4}{x^2-1}+\frac{x^6}{x^3-1}+\cdots+\frac{x^{4020}}{x^{2010}-1}$. Its ground-truth answer is $1004$.}

\medskip
\revise{\textbf{Composed question $q_{1:2}$:} An air-cored coil has a flux density of $25\,\mathrm{mT}$. When an iron core is inserted, the flux density increases to $1.6\,\mathrm{T}$. Calculate the relative permeability $\mu_r$ of the iron, and let $\kappa$ be $63$ less than $\mu_r$. Let $x$ be a complex number such that $x^{2011}=\kappa$ and $x\neq\kappa$. Compute
$\frac{x^2}{x-\kappa}+\frac{x^4}{x^2-\kappa}+\frac{x^6}{x^3-\kappa}+\cdots+\frac{x^{4020}}{x^{2010}-\kappa}$. Here $\mu_r=64$ and $\kappa=1$, so the composed prompt preserves the ground-truth answer $1004$.}
\end{tcolorbox}

\subsection{Prompts of {\datamethod}}\label{subapp:compose_prompt}

\revise{Following \cite{xiao2025ncsp}, we provide the prompt used to modify $q_1$ in \ref{appfig:p1_prompt} and the self-verification prompt used to check the modification in \ref{appfig:p1_verification}.  We use similar prompts for the other steps of \datamethod{}, and the complete prompt set is available in our released code.}

\begin{figure}[htbp!]
\centering
\begin{tcolorbox}[
  title=\textbf{Prompt for Modifying $q_1$ with $gt_1$},
  coltitle=white,
  width=\linewidth,
  halign title=center,
  left=3pt,right=3pt,top=2pt,bottom=2pt
]
Given a math problem and the final answer, your task is to find out one number from the answer and provide the corresponding definition. Follow the steps below:

Step 1: Identify a specific integer, float, or fraction within \texttt{final\_answer} and name it as $new\_variable1$; There are several situations:

\begin{enumerate}
\item If the \texttt{final\_answer} contains unknown variables:
  \begin{enumerate}
  \item If the \texttt{final\_answer} is an expression, choose one coefficient as $new\_variable1$, for example, $2x+3$, you can choose the coefficient of $x$ as $new\_variable1$, which is $2$, and in the case of $\sin(x)$, there is a hidden coefficient $1$ and a hidden amplitude $1$, you can choose either one as $new\_variable1$;
  \item If the \texttt{final\_answer} is an equation, you can choose one solution as $new\_variable1$, for example, $y=2x+1$, you can define the value of $y$ as $new\_variable1$ when given $x=1$, which is $3$;
  \item If the \texttt{final\_answer} is a symbol of an option or a word, such as `A', `B', `CAT', etc., use their first letter's order in the alphabet as a variable, such as `A' = 1, `B' = 2, `CAT' = 3, etc.;
  \item If the \texttt{final\_answer} contains 2 or more items, e.g., multiple choice questions, choose the smallest or the largest one, and then apply the corresponding situation.
  \end{enumerate}

\item If the \texttt{final\_answer} has no unknown variables, there are several situations:
  \begin{enumerate}
  \item If the \texttt{final\_answer} itself is a numerical value, like `four', `4', `$2+\sqrt{2}$', `$3\pi$', and `$\frac{3}{4}$', use it directly as $new\_variable1$;
  \item If the \texttt{final\_answer} contains 2 or more numerical values, use the largest or the smallest one as $new\_variable1$;
  \item If the \texttt{final\_answer} is an interval or ratio, choose one boundary and $\infty$ is not allowed, for example, $[2,\infty)$, you can define the lower bound as $new\_variable1$, which is $2$;
  \item If the \texttt{final\_answer} is a ratio, choose one part of the ratio, for example, $3:4$; you can define the first part of the simplified ratio as $new\_variable1$, which is $3$;
  \item If the \texttt{final\_answer} is a non-base 10 number, for example, $1001_{2}$, you can define `the number of digits in the base 2 representation' as $new\_variable1$, which is $4$;
  \item If the \texttt{final\_answer} is an angle or degree, choose the corresponding radian value, for example, $30^\circ$ or $30^\circ$, define the corresponding radian value of final answer as $new\_variable1$, which is $\pi/6$.
  \end{enumerate}
\end{enumerate}

All in all, find a way to identify a specific numerical value as $new\_variable1$ without unknown, and make sure the reader can get the value of $new\_variable1$ from the \texttt{final\_answer} through your definition.

Step 2: Output the value of $new\_variable1$, keep the exact value or math symbol, and simplify the fraction if necessary, for example, keep $\pi$ as $\pi$, keep $\sqrt{2}$ as $\sqrt{2}$, and simplify $\frac{6}{8}$ as $\frac{3}{4}$, without rounding to a decimal point.

Step 3: Output the definition of $new\_variable1$ without mentioning the real value.

\medskip
\noindent ---\par
\medskip

Output Format:
... (omit for simplicity)

Examples:
... (omit for simplicity)

\end{tcolorbox}
\caption{The Prompt for Generating Variable $v_1$ and Definition $d_1$ for $q_1$.}
\label{appfig:p1_prompt}
\end{figure}

\begin{figure}[htbp!]
\centering

\begin{tcolorbox}[
  title=\textbf{Prompt for Verifying the Modification of $q_1$},
  coltitle=white,
  width=\linewidth,
  halign title=center,
  left=3pt,right=3pt,top=2pt,bottom=2pt
]
\textbf{1. Check the extraction of variable $v_1$:}

\verb|{Problem 1}|

Assume that the final answer of the problem is \verb|{FINAL_ANSWER}|.
\verb|{DEFINITION_OF_NEW_VARIABLE1}|

Then what is the value of $new\_variable1$?

Please output the value of $new\_variable1$ directly, wrapping it in \verb|\boxed{}|, for example, \verb|\boxed{3}|.
\newline

\textbf{2. Check the value of $v_1$ using Python:}

**Task Description:**

Write a Python program to compare two given values and determine if they are equal.
Follow these guidelines:

\begin{enumerate}
  \item Use the \texttt{sympy} library to handle symbolic comparisons, ensuring that
        equivalent expressions (e.g., $\frac{2}{4}$ and $\frac{1}{2}$) are recognized as equal.
  \item For values involving irrational constants (e.g., $\pi$, $e$), perform comparisons up to
        \textbf{two decimal places} for practical equivalence.
  \item Include clear intermediate steps in the program, such as evaluating or simplifying the
        values where appropriate.
  \item Wrap the final comparison outcome in a \verb|\boxed{}| command for clarity.
  \item Provide both the Python code and the results of running the code.
\end{enumerate}

**Output Format:**

\begin{verbatim}
```python
{The Python code that compares the two given values, including print statements
for intermediate steps and the \boxed{final comparison outcome}.}
```

```output
{The output of the Python program.}
```

---

[Examples Here]
\end{verbatim}
\end{tcolorbox}

\caption{The Prompt for Verifying the Modification of $q_1$.}
\label{appfig:p1_verification}
\end{figure}

\clearpage
\newpage
\section*{NeurIPS Paper Checklist}

\begin{enumerate}

\item {\bf Claims}
    \item[] Question: Do the main claims made in the abstract and introduction accurately reflect the paper's contributions and scope?
    \item[] Answer: \answerYes{} %
    \item[] Justification: Our abstract and introduction summarize our scope and contribution. %

\item {\bf Limitations}
    \item[] Question: Does the paper discuss the limitations of the work performed by the authors?
    \item[] Answer: \answerYes{} %
    \item[] Justification: We include a Limitation Section at the beginning of the Appendix. %

\item {\bf Theory assumptions and proofs}
    \item[] Question: For each theoretical result, does the paper provide the full set of assumptions and a complete (and correct) proof?
    \item[] Answer: \answerNA{}%
    \item[] Justification: This paper does not include theoretical results.  %

    \item {\bf Experimental result reproducibility}
    \item[] Question: Does the paper fully disclose all the information needed to reproduce the main experimental results of the paper to the extent that it affects the main claims and/or conclusions of the paper (regardless of whether the code and data are provided or not)?
    \item[] Answer: \answerYes{} %
    \item[] Justification: We clearly disclose our experimental setting in~\cref{subsec:exp_setup} and~\cref{app:exp_detail}. %

\item {\bf Open access to data and code}
    \item[] Question: Does the paper provide open access to the data and code, with sufficient instructions to faithfully reproduce the main experimental results, as described in supplemental material?
    \item[] Answer: \revise{\answerYes{}} %
    \item[] Justification: \revise{Our code, data, and models are publicly available.} %

\item {\bf Experimental setting/details}
    \item[] Question: Does the paper specify all the training and test details (e.g., data splits, hyperparameters, how they were chosen, type of optimizer) necessary to understand the results?
    \item[] Answer: \answerYes{} %
    \item[] Justification: We clearly disclose our experimental setting in~\cref{subsec:exp_setup} and~\cref{app:exp_detail}. %

\item {\bf Experiment statistical significance}
    \item[] Question: Does the paper report error bars suitably and correctly defined or other appropriate information about the statistical significance of the experiments?
    \item[] Answer: \answerNo{} %
    \item[] Justification: As stated in~\cref{subsec:exp_setup} and~\cref{app:exp_detail}, we follow the standard recipe to report \texttt{pass@1}.

\item {\bf Experiments compute resources}
    \item[] Question: For each experiment, does the paper provide sufficient information on the computer resources (type of compute workers, memory, time of execution) needed to reproduce the experiments?
    \item[] Answer: \answerYes{} %
    \item[] Justification: Partially yes, all the training details are given in~\cref{subapp:training}.%
    
\item {\bf Code of ethics}
    \item[] Question: Does the research conducted in the paper conform, in every respect, with the NeurIPS Code of Ethics \url{https://neurips.cc/public/EthicsGuidelines}?
    \item[] Answer:  \answerYes{} %
    \item[] Justification: We adhere to the NeurIPS Code of Ethics.%

\item {\bf Broader impacts}
    \item[] Question: Does the paper discuss both potential positive societal impacts and negative societal impacts of the work performed?
    \item[] Answer: \answerYes{} %
    \item[] Justification: We have included an ``Impact Statement'' section in the appendix.
    
\item {\bf Safeguards}
    \item[] Question: Does the paper describe safeguards that have been put in place for responsible release of data or models that have a high risk for misuse (e.g., pre-trained language models, image generators, or scraped datasets)?
    \item[] Answer: \answerNA{} %
    \item[] Justification: This paper poses no such risks

\item {\bf Licenses for existing assets}
    \item[] Question: Are the creators or original owners of assets (e.g., code, data, models), used in the paper, properly credited and are the license and terms of use explicitly mentioned and properly respected?
    \item[] Answer: \answerYes{} %
    \item[] Justification: We clearly cited all existing assets in~\cref{subsec:exp_setup} and~\cref{app:exp_detail}.

\item {\bf New assets}
    \item[] Question: Are new assets introduced in the paper well documented and is the documentation provided alongside the assets?
    \item[] Answer: \answerYes{} %
    \item[] Justification: The details of generating new assets are given in~\cref{app:data_composition}.

\item {\bf Crowdsourcing and research with human subjects}
    \item[] Question: For crowdsourcing experiments and research with human subjects, does the paper include the full text of instructions given to participants and screenshots, if applicable, as well as details about compensation (if any)? 
    \item[] Answer: \answerNA{} %
    \item[] Justification: This paper does not involve crowdsourcing nor research with human subjects.

\item {\bf Institutional review board (IRB) approvals or equivalent for research with human subjects}
    \item[] Question: Does the paper describe potential risks incurred by study participants, whether such risks were disclosed to the subjects, and whether Institutional Review Board (IRB) approvals (or an equivalent approval/review based on the requirements of your country or institution) were obtained?
    \item[] Answer: \answerNA{} %
    \item[] Justification: This paper does not involve crowdsourcing nor research with human subjects.

\item {\bf Declaration of LLM usage}
    \item[] Question: Does the paper describe the usage of LLMs if it is an important, original, or non-standard component of the core methods in this research? Note that if the LLM is used only for writing, editing, or formatting purposes and does \emph{not} impact the core methodology, scientific rigor, or originality of the research, declaration is not required.
    \item[] Answer: \answerYes{} %
    \item[] Justification: We include a ``Declaration of LLM Usage'' section to discuss LLM use.

\end{enumerate}

\end{document}